# StomaD²: An All-in-One System for Intelligent Stomatal Phenotype Analysis via Diffusion-Based Restoration Detection Network

*Quanling Zhao[1,\*], Meng'en Qin[1,\*], Yanfeng Sun[3], Yuan Miao[2,†], Xiaohui Yang[1,†]*

[1]Henan Engineering Research Center for Artificial Intelligence Theory and Algorithms, School of Mathematics and Statistics, Henan University, Kaifeng 475004, China.

[2]International Joint Research Laboratory for Global Change Ecology, School of Life Sciences, Henan University, Kaifeng 475004, China.

[3]State Key Laboratory of Cotton Biology, School of Life Sciences, Henan University, Kaifeng 475004, China.

\* Equal contribution. † Corresponding author.

**Abstract:** Stomata play a crucial role in regulating plant physiological processes and reflecting environmental responses. However, accurate and high-throughput stomatal phenotyping remains challenging, as conventional approaches rely on destructive sampling and manual annotation, restricting large-scale and field deployment. To overcome these limitations, a noninvasive restoration-detection integrated framework, termed StomaD², is developed to achieve accurate and fast stomatal phenotyping under complex imaging conditions. The framework incorporates a diffusion-based restoration module to recover degraded images and a specialized rotated object detection network tailored to the small, dense, and cluttered characteristics of stomata. The proposed network enhances feature representation through three key innovations: a column-wise structure for global feature interaction, context-aware resampling and reweighting mechanism to improve multi-scale consistency, and a feature reassembly module to boost discrimination against complex backgrounds. In extensive comparisons, StomaD² demonstrated state-of-the-art performance. On public Maize and Wheat datasets, it achieved accuracies of 0.994 and 0.992, respectively, significantly outperforming existing benchmarks. When benchmarked against ten other advanced models, including Oriented Former and YOLOv12, StomaD² achieved a top-tier F1-score/mAP of 0.989. The framework is integrated into a user-friendly, field-operable system that supports the fast extraction of eight stomatal phenotypes, such as density and conductance. Validated on more than 130 plant species, StomaD²'s results highlight its strong generalizability and potential for large-scale phenotyping, plant physiology analysis, and precision agriculture applications.

## 1 Introduction

Stomata (Fig. 1) are channels on the surface of plant leaves [1]. Open stomatal pores allow the absorption of carbon dioxide into the plant for photosynthesis, while simultaneously allowing oxygen and water vapor to diffuse out [2]. Stomata play a crucial role in determining crop productivity by regulating photosynthesis and water utilization, both of which depend on stomatal conductance [3]. Given the impact of stomatal phenotypes, such as shape, density, area, and distribution on crop productivity across diverse environmental conditions [4], genetic variation in these phenotypes has emerged as a crucial focus for crop improvement efforts [7]. The most direct assessment of drought tolerance involves evaluating water-use efficiency and yield performance under water-limited and other biotic stress conditions [5]. Since stomatal density on leaves is closely associated with a plant's water-

---

The code is given (GitHub - dear13-star/StomaD2).

use efficiency [6], its measurement can provide valuable insights for identifying stress-resilient genes, thereby improved breeding strategies. Stomatal behavior including the dynamic opening and closing regulated by guard cell turgor changes [8]. Therefore, accurately acquiring high-throughput stomatal phenotypic data is particularly important for researchers.

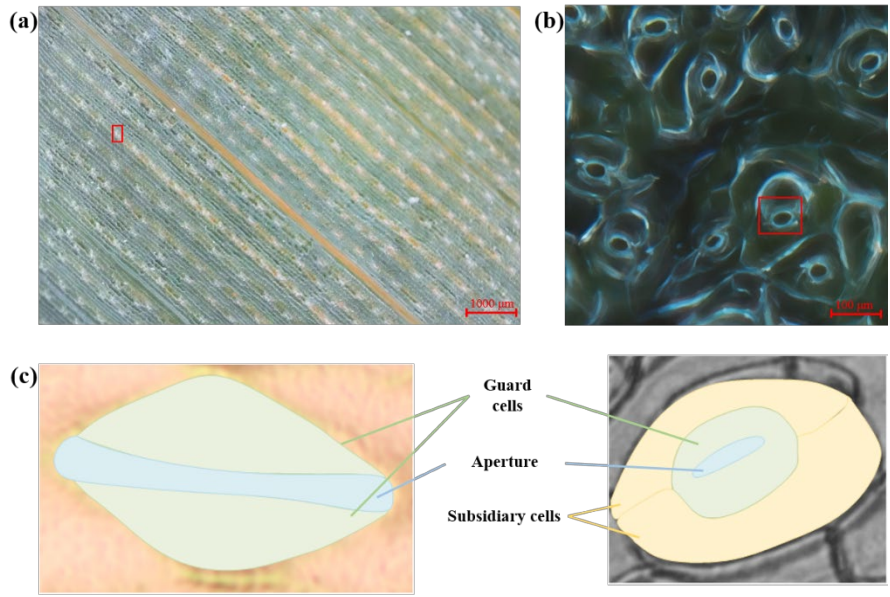

**Fig. 1** Stomata of maize and peanut. Maize images were captured at 2736 × 1824 pixel resolution, and peanut images were captured at 640 × 640 pixel resolution with handheld microscope. Components of maize and peanut stomata (c) that are rotated and magnified from original images (a) and (b) (indicated by red bounding box) are highlighted in pseudo-color and labelled; scale bars, 100 μm.

Due to the importance of stomata, investigating stomatal regulation has become a common task for biologists studying plants [9]. To study stomatal phenotype (density, conductance, etc.) and behavior, researchers typically require the acquisition of stomatal images. Traditional methods for obtaining stomatal phenotypic data often rely on destructive sampling techniques, such as leaf punching, detachment, or chemical treatment to capture information at a single time point, typically through manual observation and analysis [10]. However, this is a tedious, error-prone, time-consuming and, in some cases, subjective task due to the large number of stomata in each image.

Recent efforts have improved these traditional methods. Bourdais et al. [11] achieved high-efficiency imaging using a spinning-disk automated Opera microscope for leaves on 96-well plates. Millstead et al. [12] mounted impressions on microscope slides to enhance the conventional nail polish method, reducing unevenness and improving focus depth. With the impressive performance of deep learning methods in interdisciplinary researches [47][48], an increasing number of people are using them to assist in plant stomatal phenotypic analysis. Kwong et al. [14] in order to accelerate the selection process for drought-resistant breeding in oil palm, used MobileNet as a template to automatically detect stomata in oil palm leaves, thereby determining the stomatal density of drought-resistant oil palm varieties. Solimani et al. [16] developed a tool based on the YOLOv8 for plant stomata detection and phenotype data collection. Fetter et al. [17] introduce StomataCounter, an automated stomata counting system using a deep convolutional neural network to identify stomata in a variety of different microscopic images. Zhang et al. [18] proposed a deep learning-based rotated stomata detection method for detecting damaged maize stomata, enabling the extraction of stomatal conductance information. Gu et al. [19]

developed an automatic tomato fruit detection and phenotypic trait extraction method based on an improved RTDETR model, enabling accurate, rapid, and nondestructive measurement of fruit characteristics to support robotic harvesting, yield prediction, and variety breeding. Yang et al. [15] proposed SLPA-Net, which improves stomatal localization accuracy by optimizing the loss function. [20] proposed RotatedStomataNet, which treats stomata as rotated detection targets and enables real-time stomatal phenotypic analysis. However, these approaches predominantly rely on destructive sampling. As crop science increasingly advances toward practical and industrial applications, it is imperative to develop noninvasive methods for stomatal phenotyping suitable for field conditions.

Another approach for stomatal imaging is noninvasive. Noninvasive technologies enable continuous and periodic monitoring of leaf morphology without interfering with plant growth. This facilitates the study of dynamic physiological processes such as leaf lifecycle progression, changes in photosynthetic capacity, and developmental rhythms. Moreover, by avoiding tissue damage and physiological disturbances typically caused by traditional methods, these techniques yield experimental results that more closely reflect natural conditions. Sun et al. [13] acquired live leaf images using a portable microscope, preserving the plant's vitality, thus improving imaging efficiency. Liang et al. [21] proposed StomataScorer, a new method for detecting stomata and extracting stomatal traits, the method utilizes images of living stomata in maize leaves. Song et al. [22] proposed an automatic stomatal segmentation and parameter calculation method for plant leaf microscopic images based on Mask R-CNN. Yang et al. [33] proposed an automated phenotypic analysis framework for strawberries, integrating mobile and robotic image acquisition with an improved semantic segmentation model to enable accurate organ segmentation and germplasm classification across different growth stages. However, the acquired images may suffer from quality instability, such as blurriness or low quality, which will greatly impact the accuracy of the model in subsequent stomatal phenotype data acquisition. Moreover, most existing methods are primarily designed for monocotyledonous plants, whose stomata are typically arranged in a relatively orderly manner. In contrast, stomata in dicotyledonous plants tend to be smaller, more densely packed, and irregularly distributed, which significantly increases the complexity and difficulty of noninvasive detection.

To address these challenges, we propose StomaD$^2$, a novel method for accurate and high-throughput stomatal phenotyping from images or videos captured by handheld microscopes even when blurred. StomaD$^2$ automatically identifies stomata in both monocotyledonous and dicotyledonous leaves and extracts key phenotypic traits such as stomatal density, aperture density, conductance et al. The main contributions of this work are as follows:

- In this paper, we employ a diffusion model to address the issue of blurred noninvasive stomatal images caused by imaging or transmission limitations. This approach enhances the clarity and quality of stomatal features, thereby facilitating more accurate acquisition of physiological phenotypic data.
- To address the small, dense, and cluster of noninvasive stomata, we propose a rotated object detection network. This design disentangles stomatal features at each scale by integrating multi-level information, enhancing intra-scale discrimination while suppressing background noise and artifacts, thus achieving more accurate and robust stomatal localization.
- Finally, we present a user-friendly software, StomaD$^2$, which integrates image restoration and

stomatal detection into a unified framework. It enables one-stop automatic collection of stomatal phenotypic data, eliminating the need for manual, step-by-step operations and streamlining the entire analysis process.

## 2 Materials and Method

### 2.1 Plant Materials

The crops were cultivated at the He Yuan Plant Cultivation Base in Kaifeng and the Xinxiang Comprehensive Experimental Station of the Chinese Academy of Agricultural Sciences, while the image data were provided by the State Key Laboratory of Cotton Biology and the State Key Laboratory of Crop Stress Adaptation and Improvement at Henan University. Damaged stomatal images were captured using a ToupCam S500-GS optical microscope, with maize and Arabidopsis image resolutions of 2736×1824 and 2448×2048, respectively. The noninvasive dataset was acquired using a ProScope HR5 to image living leaves, with resolutions of 2592×1944 and 1920×1440, where 224 pixels correspond to 100 μm. StomaD$^2$ employs two types of datasets (Fig. 2): noninvasive stomatal images captured under natural field conditions and destructive images obtained in a controlled laboratory environment.

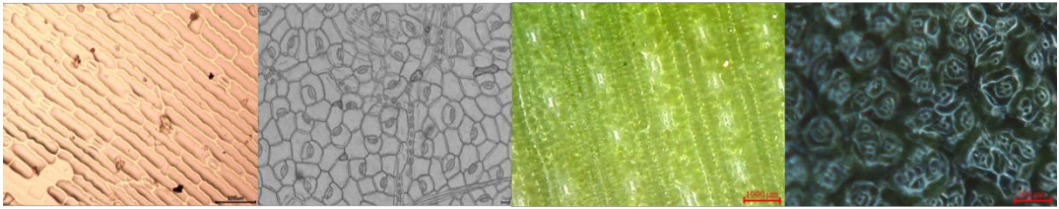

**Fig. 2** Examples of stomatal images acquired by optical and handheld microscopes. From left to right: destructive maize and peanut stomata, noninvasive maize and peanut stomata.

In addition to the inner datasets, two publicly available stomatal datasets were included to enhance benchmarking and cross-species evaluation. First, high-quality microscopic images of maize and wheat were obtained from StomataHub, which provides standardized stomatal datasets for algorithm development and comparative analysis. Second, to assess large-scale generalizability, we further incorporated a broad stomatal image collection from the United States National Museum (USNM), comprising over 130 plant species across angiosperms and gymnosperms. This multi-source dataset enables rigorous evaluation of the robustness, adaptability, and cross-species transferability of StomaD².

### 2.2 Data Annotation

**Table 1** The plant species and their quantities used for training and testing

| Types | | Optical microscope | | Handled microscope | |
|---|---|---|---|---|---|
| | | Train | Test | Train | Test |
| dicotyledonous | soybean | 500 | 50 | 0 | 0 |
| | begonia | 1000 | 50 | 0 | 0 |
| | peanut | 500 | 50 | 3000 | 200 |
| | broad bean | 0 | 0 | 1000 | 50 |
| monocotyledonous | Arabidopsis | 2000 | 100 | 0 | 0 |
| | maize | 1000 | 100 | 100 | 20 |

All images were manually annotated using RolabelImg, generating XML files with rotated bounding boxes labeled as either "stoma" or "aperture." To improve the model's ability to learn morphological features, data augmentation techniques such as random flipping and mosaic were applied to a subset of the dataset. The model was then trained and validated on the augmented data (Table 1).

**2.3 Diffusion-Based Restoration Detection Network**

This section introduces the proposed method, including the restoration detection pipeline for stomatal image enhancement and detection. The detailed structures are described below. The full workflow of StomaD² is illustrated in Fig. 3.

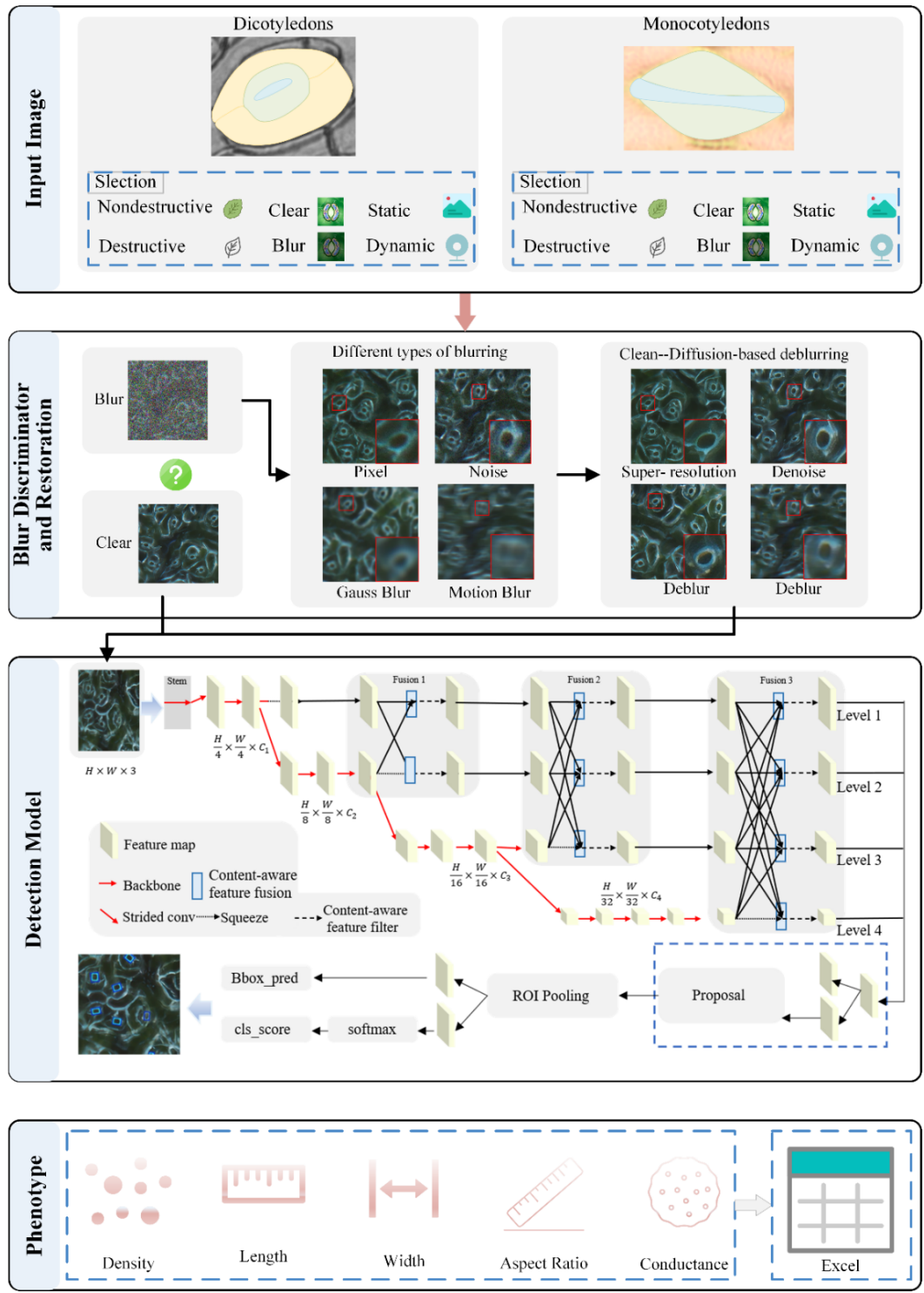

**Fig. 3** Illustrates the workflow of StomaD$^2$(take peanut as an example), which consists of four steps: data input, image quality assessment, detection models and phenotype. The data includes noninvasive images captured in a completely natural environment and destructive images obtained in a laboratory setting. The model is divided into two parts: the restoration model and the detection model. If an image is determined to be blurry, it is processed by the restoration detection model; if the image is deemed clear, it is input directly into the detection model and access phenotype data.

**Stomata Restoration:** Since high-quality images are essential for accurate phenotype acquisition, we focus on improving image quality. In this work, we adopt a two-stage enhancement strategy inspired by [23], where a regression network first restores the coarse stoma structure and a diffusion model then refines stoma details.

**Stage One**: We adopt SwinIR [25] pretrained on ImageNet [24] as the initial model. Deep features are extracted using multiple Residual Swin Transformer Blocks, each comprising several Swin Transformer Layers. Shallow and deep features are fused to preserve both low and high-frequency information. The model is trained using $L_2$ pixel loss. The formula for the loss function is as follows:

$$I_{reg} = SwinIR(I_{LQ}), \mathcal{L}_{RM} = I_{reg} - I_{HQ}, \tag{1}$$

where ($I_{HQ}$) and ($I_{LQ}$) represent the high-quality and low-quality corresponding images, respectively, and ($I_{reg}$) is the image obtained through regression learning, which will be used for fine-tuning the latent diffusion model.

**Stage Two:** Although the regression stage can recover the major structure of the stoma, residual degradation may still remain, especially in regions with severe blur or noise contamination. Therefore, a diffusion-based refinement stage is further introduced to suppress residual degradation while preserving local stomatal structures. We use the fine-tuned Stable Diffusion [26] to reconstruct the image with the obtained $(I_{reg} - I_{HQ})$. The VAE encoder [27] maps $(I_{reg})$ into the latent space, resulting in the conditional latent $(\varepsilon(I_{reg}))$. Objective function is as follow:

$$\mathcal{L}_{Diff} = \mathbb{E}_{z_t, c, t, \varepsilon, \varepsilon(I_{reg})} \left[ \left\| \varepsilon - \varepsilon_\theta \left( z_t, c, t, \varepsilon, \varepsilon(I_{reg}) \right) \right\|_2^2 \right], \tag{2}$$

where $\varepsilon \sim \mathcal{N}(0, I)$, $z_t$ is the latent code, $c$ is the sample point, and $t$ represents the time step, which is uniformly sampled.

Overall, the enhancement module serves as a preprocessing component, thereby improving localization robustness without treating restoration as an independent objective.

**Stomata Detection**: To obtain more accurate phenotypic data, we design a progressive content-aware feature refinement pyramid network (Fig. 3). It employs a horizontally-spread column-wise network that starts by fusing adjacent-level features and progressively achieves non-adjacent and global feature interaction [42]. Lightweight CSPDarknet [28] is used as the backbone to extract both stoma texture and semantic features from stoma images. During fusion, we propose the content-aware feature fusion and feature filter modules.

**Content-Aware Feature Fusion**: Due to the repetitive convolution and pooling operations in the backbone, there still exists a predictable spatial position discrepancy between the sampled feature maps $(f_i^s)$ and unsampled feature maps $(f_k^u)$, as shown in **Fig. 4**(a). Before performing feature interaction, it is necessary to adjust $(f_i^s)$ based on the spatial position information provided by $(f_k^u)$. We achieve this by learning an offset matrix (location content), where each value can be interpreted as the offset in the 2D space between corresponding points in $(f_i^s)$ and $(f_k^u)$. The adjusted feature $f_i^a$ can be expressed as:

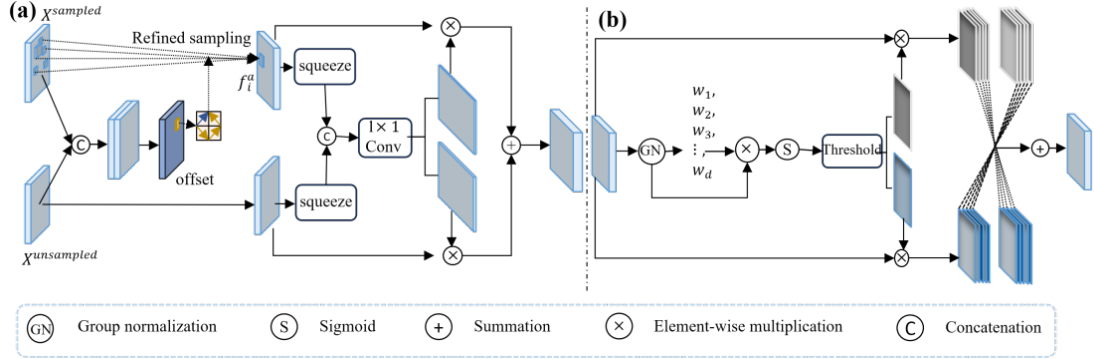

**Fig. 4** The architecture of content-aware feature fusion and content-aware feature filter. (a) represents the content-aware feature fusion with refined sampling (b) represents the content-aware feature filter.

$$f_i^a = f_{adjust}(f_i^s, f_\Delta(f_i^s, f_k^u)), \quad (3)$$

where $f_i^s$ is the $i$-th layer sampled feature map and $f_k^u$ is the $k$-th layer unsampled feature map. $\Delta f$ represents the offset learned from the spatial difference, and $f_{adjust}(\cdot)$ denotes the operation of adjusting the coarsely sampled feature map based on the learned offset values.

Let $f_{i \to k}^a$ represent the feature aligned and adjusted from layer $i$ to layer $k$, where the feature of layer $k$, $f^k$ is given by:

$$f^k = \sum_{i=1}^{s} w^i f_{i \to k}^a, \quad (4)$$

where $w^i$ denotes the spatial weight of the feature map of layer $i$ and is shared across all channels, $f_{i \to k}^a = f_k^u$ if $i = k$. For each $w_{mn}^i$ in $w^i$, we have $w_{mn}^i \in [0,1]$.

**Content-Aware Feature Filter:** To further refine multi-layer features, we introduce the content-aware feature filter module, as shown in Fig. 4(b). It further facilitates intra-scale discriminative features and filters out redundant planting background and dirt features for more accurate and quality stoma locating based on information content and spatial variation of feature maps.

For the fused feature $f^k$, it is necessary to separate the feature channels with high information content along the spatial dimensions within the layer, and reconstruct the feature channels with low information content. This further enhances the effect of multi-scale feature fusion. Group Normalization (GN) uses learnable scaling factors to assess the information content of feature maps. For $f \in \mathbb{R}^{N \times C \times H \times W}$ (where $H$ and $W$ represent height and width, $C$ represents the number of channels, and $N$ is the batch size), group normalization is defined as follows:

$$f_{std} = GN(X) = \alpha \frac{f - \mu}{\sqrt{\sigma^2}}, \quad (5)$$

where $\sigma$ and $\mu$ are the standard deviation and mean of $f$, respectively, and $\alpha$ and $\beta$ are learnable coefficients. The greater the spatial difference, the richer the spatial information, indicating that the corresponding feature contains more informative content. The weights are derived from the following formula, which represents the importance of the features:

$$w = \frac{\alpha_i}{\sum_{j=1}^{C} \alpha_j}, i, j = 1, 2, \cdots, C, \quad (6)$$

The feature channels are re-weighted spatially using $w$, and then the weighted values are transformed into the range (0,1) using Sigmoid function to compute the final weights. Subsequently, weights above a certain threshold are set to 1 to generate the information weight $w_1$, while weights below the threshold are set to 0 to generate the non-informative weight $w_2$. The calculation of $w_1$ and $w_2$ can be expressed as:

$$w_1 = \begin{cases} Sigmoid(w * f_{std}), & Sigmoid(w * f_{std}) \leq threshold, \\ 1, & Sigmoid > threshold, \end{cases} \quad (7)$$

$$w_2 = \begin{cases} Sigmoid(w * f_{std}), & Sigmoid(w * f_{std}) \geq threshold, \\ 0, & Sigmoid < threshold. \end{cases} \quad (8)$$

The fused feature $f^k$ is then multiplied by $w_1$ and $w_2$ to obtain the features with high information content $f^{w_1}$ and low information content $f^{w_2}$, respectively. Finally, we add $f^{w_1}$ and the reversed $f^{w_2}$ together to form the refined feature.

**2.4 Maximum Conductance Calculation**

Anatomical $Gs_{max}$ often exceeds operational gs by several fold [31], but works in parallel with gs at a spatial and temporal scale to optimise stomatal responses to the prevailing environmental conditions [32]. High $Gs_{max}$ precludes high gs under yield potential conditions and can be used to predict gs under well-watered, light-saturated environments [32]. A measurement of stomatal size allows a calculation of the potential maximal rate of $G_s$ to water vapor, known as anatomical maximum stomatal conductance:

$$Gs_{max} = \frac{f \times D_{stoma} \times \gamma_{max}}{1.6v \left( w + \frac{\pi}{2}\sqrt{\frac{\gamma_{max}}{\pi}} \right)}, \quad (9)$$

$$\gamma_{max} = \pi \times l_{stoma}^2 / 4, \quad (10)$$

where $f$ represents the diffusion coefficient of water in air, which is ideally set to $(2.49 \times 10^{-5}\ m^2/s)$. $D_{stoma}$ denotes the stoma density (the number of stomata per square millimeter), $\gamma_{max}$ represents the mean stoma opening area, $v$ is the molar volume of air, which is 22.4 $L/mol$ at 25°C and 101.3 kPa. $w$ denotes the width of the guard cells, and $l_{stoma}$ represents the stoma length, and $\pi$ is the mathematical constant taken as 3.142 [34].

**3 Experiments and Results**

**3.1 Experiment Setting and Evaluation Metrics**

Training employs stochastic gradient descent with an initial learning rate of 0.02, momentum of 0.9, and weight decay of 0.0001. To boost detection performance, images are cropped to multiple sizes (640×640, 1024×1024, and 1920×1440). Experiments are conducted on a desktop with a 256GB SSD, Intel Xeon Gold 5218R CPU (2.10GHz), and an RTX 4060 GPU.

To comprehensively evaluate the performance of StomaD[2], we use quantitative metrics. For detection performance, object detection metrics are used, including Precision, Recall, F1 score, Average Precision (AP), and mean Average Precision (mAP), all of which indicate better performance when values are higher. To assess the agreement between the predicted and manually measured phenotypic data, we use the Concordance Correlation Coefficient (CCC), which ranges from −1 to 1, with values closer to 1 indicating stronger consistency. Given that stomata and their apertures are generally elliptical, we define stomatal length and width, respectively. Beyond detection and counting accuracy, we further evaluate the quality of phenotypic estimation using metrics such as length accuracy, width accuracy, average length accuracy, and average width accuracy, where higher values denote better performance. Additionally, we report mean square error (MSE) and root mean square error (RMSE), both of which should be as low as possible, where smaller differences from ground truth indicate better accuracy.

In addition to detection and phenotypic estimation, we also assess image quality before and after restoration. Image quality evaluation is based on the distribution tail properties of image spatial frequencies, characterized by the standard deviation (fSTD) and mean value (fMean) of the frequency tail. Low fSTD and fMean values indicate image blur, while high values suggest sharper, clearer images. A third metric, tEntropy, measures the information content of the image—higher values represent high-contrast or noisy images, whereas lower values indicate low contrast. These image quality metrics are calculated using PyImq [35] and are normalized between 0 and 1. (Detailed definitions and formulas are provided in the Supporting Information).

## 3.2 Evaluation of Stoma Restoration and Detection Performance

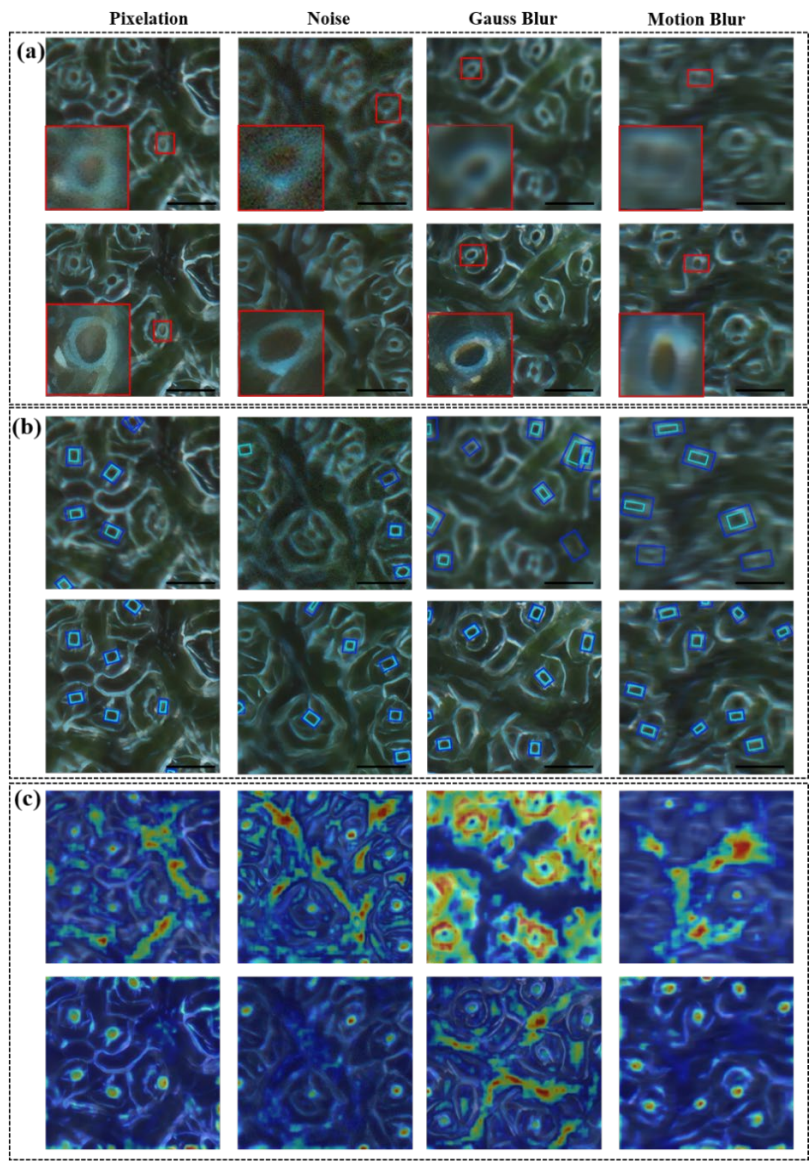

**Fig. 5** Different types of blurred images, their restored results, and corresponding visualizations. (a) Local enlargements of blurred and restored regions. (b) Detection outputs on original and restored images. (c) Heatmaps illustrating feature activations before and after restoration.

As low-quality images can adversely affect downstream tasks, StomaD² incorporates restoration for four common degradation types to enhance detection performance. Fig. 5 provides a qualitative comparison of the restoration results: Fig. 5(a) shows images before and after restoration, with red boxes marking magnified regions, Fig. 5(b) presents the corresponding detection results and Fig. 5(c) displays heatmaps of stomatal energy. The restored images exhibit clearer structures and more concentrated feature responses, suggesting improved input quality and enhanced detection capability.

Table 2 demonstrates that the proposed restoration method effectively improves image quality across various degradation types. In most cases, fMean, fSTD and tEntropy increase significantly after restoration, indicating enhanced structural and textural fidelity. Notably, for noise degradation, although these metrics decrease, the reduction is expected due to the removal of high-frequency noise components, suggesting successful denoising rather than detail loss. reports image quality metrics before and after restoration.

Table 2 Quantitative metrics of image quality before and after restoration

|  |  | fMean | fSTD | tEntropy |
|---|---|---|---|---|
| Before Restoration | pixelation | 151306.845 | 203543.061 | 6.446 |
| After Restoration |  | 10289674.492 | 14935723.300 | 6.447 |
| Before Restoration | Noise | 72475540.800 | 72255318.400 | 6.782 |
| After Restoration |  | 11251529.875 | 18393064.683 | 6.449 |
| Before Restoration | Motion Blur | 33160.540 | 33099.093 | 6.232 |
| After Restoration |  | 45634.195 | 110291.510 | 6.436 |
| Before Restoration | Gauss Blur | 36802.103 | 36702.031 | 6.340 |
| After Restoration |  | 13898813.700 | 22409713.667 | 6.484 |

Table 3 Detection performance metrics before and after restoration

|  |  | pixelation | Noise | Motion Blur | Gauss Blur |
|---|---|---|---|---|---|
| Before Restoration | Stoma AP | 0.654 | 0.349 | 0.004 | 0.261 |
| After Restoration |  | 0.884 | 0.770 | 0.234 | 0.845 |
| Before Restoration | Aperture AP | 0.607 | 0.380 | 0.005 | 0.307 |
| After Restoration |  | 0.866 | 0.839 | 0.232 | 0.819 |
| Before Restoration | mAP | 0.701 | 0.364 | 0.003 | 0.215 |
| After Restoration |  | 0.903 | 0.804 | 0.236 | 0.871 |
| Before Restoration | Stoma F1 score | 0.501 | 0.370 | 0.120 | 0.456 |
| After Restoration |  | 0.789 | 0.786 | 0.489 | 0.820 |
| Before Restoration | Aperture F1 score | 0.868 | 0.355 | 0.015 | 0.298 |
| After Restoration | | 0.952 | 0.850 | 0.409 | 0.859 |

Table 3 summarizes the detection performance before and after image restoration under various types of degradation. Across all evaluated metrics—including Stoma AP, Aperture AP, mAP, and F1 scores—restoration consistently leads to significant improvements. This indicates that the proposed method effectively enhances structural clarity and boundary definition, which are critical for accurate stomatal and aperture detection. In particular, substantial gains are observed under common degradations such as pixelation, noise, and Gaussian blur. Even in the most challenging case of motion blur, the restored images enable markedly improved detection performance. These results suggest that the restoration framework contributes not only to enhanced image quality, but also plays a significant role in boosting the performance of downstream detection tasks.

Table 4 Objective Evaluation of StomaD² on Public Benchmark Datasets

| DataSet | Source | Method | mAP | F1 score | Citation |
|---|---|---|---|---|---|
| Maize (ProScope HR2) | StomataHub | StomaSoccer | - | 0.970 | Liang X.Y. ea al. [46] |
| | | **StomaD²** | **0.994** | **0.995** | **This paper** |
| Wheat | StomataHub | StomataTracker | 0.958 | - | Sun Z.Z. ea al. [13] |
| | | **StomaD²** | **0.992** | **0.978** | **This paper** |

Table 5 Quantitative and comparative listing of test results in terms of AP, mAP, F1 score on peanut dataset, where the best results are boldfaced.

| Methods | Stoma | Aperture | mAP | Stoma F1 score | Aperture F1 score | Citation |
|---|---|---|---|---|---|---|
| Rotated Faster R-CNN | 0.900 | 0.908 | 0.904 | 0.892 | 0.895 | S. Yang et al. [36] |
| Oriented RCNN | 0.903 | 0.908 | 0.905 | 0.901 | 0.896 | Xie et al [37] |
| R³Det | 0.896 | 0.633 | 0.765 | 0.892 | 0.705 | X. Yang et al. [38] |
| Oriented Former | 0.948 | 0.974 | 0.961 | 0.955 | 0.920 | Zhao et al. [39] |
| YOLOv8 | 0.912 | 0.992 | 0.952 | 0.927 | 0.927 | Varghese & Sambath [40] |
| YOLOv12 | 0.976 | 0.988 | 0.982 | 0.938 | 0.935 | Tian et al. [41] |
| StoManager1 | 0.974 | 0.975 | 0.974 | 0.953 | 0.956 | J.X. Wang et al. [43] |
| StomataGSMAX | 0.990 | 0.969 | 0.980 | 0.990 | 0.900 | Gibbs Jonathon A et al. [44] |
| StomaGAN | 0.998 | 0.586 | 0.793 | 0.994 | 0.588 | Gibbs Jonathon A., & Gibbs Alexandra J. [45] |
| **StomaD²** | **0.990** | **0.988** | **0.989** | **0.972** | **0.973** | **This paper** |

**Table 6** Detection results of StomaD² on our dataset

| Dataset | Precision | Recall | F1 score | AP | mAP |
|---|---|---|---|---|---|
| Peanut stomata (Dicot) | 0.976 | 0.968 | 0.972 | 0.990 | 0.989 |
| Peanut aperture (Dicot) | 0.977 | 0.970 | 0.973 | 0.988 | |
| Maize stomata (Monocot) | 0.958 | 0.970 | 0.964 | 0.987 | 0.987 |

Table 5 presents a quantitative comparison of detection performance on the Peanut dataset across various methods. The evaluation includes state-of-the-art object detection models such as YOLOv8 and YOLOv12, as well as existing approaches specifically designed for stomatal detection. StomaD² is also validated on the publicly available StomataHub dataset (Table 4). The results show that StomaD² consistently outperforms other methods in both stomatal and aperture detection tasks, indicating its strong capability to handle challenging noninvasive dicotyledonous leaf images and its broad applicability across different crop species. Table 6 presents the detection performance of stomata and apertures in Peanut (dicot) and Maize (monocot) datasets, evaluated using precision, recall, F1 score, and AP. All tasks achieved high precision and recall, with F1 scores above 0.96. The AP values reached 0.988 or higher, and the mean average precision (mAP) was 0.989 for Peanut and 0.987 for Maize. These results demonstrate the effectiveness of StomaD² in accurately detecting stomatal structures.

## 3.3 Evaluation of Measurement Accuracy and Error between Predicted and Manually Labeled Results

The violin plots in Fig. 6 show no significant difference in stomatal and aperture densities between manual annotations and model predictions ($P > 0.05$), suggesting strong consistency. In addition, Fig. 7 reports the CCC and relative errors (RE) for specific traits. For the peanut dataset, the CCCs for stomatal length, stomatal width, aperture length, and aperture width are 0.9772, 0.9800, 0.9787, and 0.9834, respectively. For the maize dataset, stomatal length and width show CCCs of 0.9779 and 0.9574, respectively. Notably, the relative errors across all traits are generally below 0.06, indicating high accuracy and low deviation from manual annotations.

Accuracy results in Table 7 and Fig. 7(c) summarizes the prediction performance of stomatal and aperture length and width on Peanut and Maize datasets. In general, the model achieved high accuracy across all tasks, with average values exceeding 96%. On the Peanut dataset, stomatal measurements yielded slightly higher accuracy, while aperture predictions exhibited lower RMSEs. For Maize, stomatal predictions maintained reliable accuracy, though with relatively higher RMSEs.

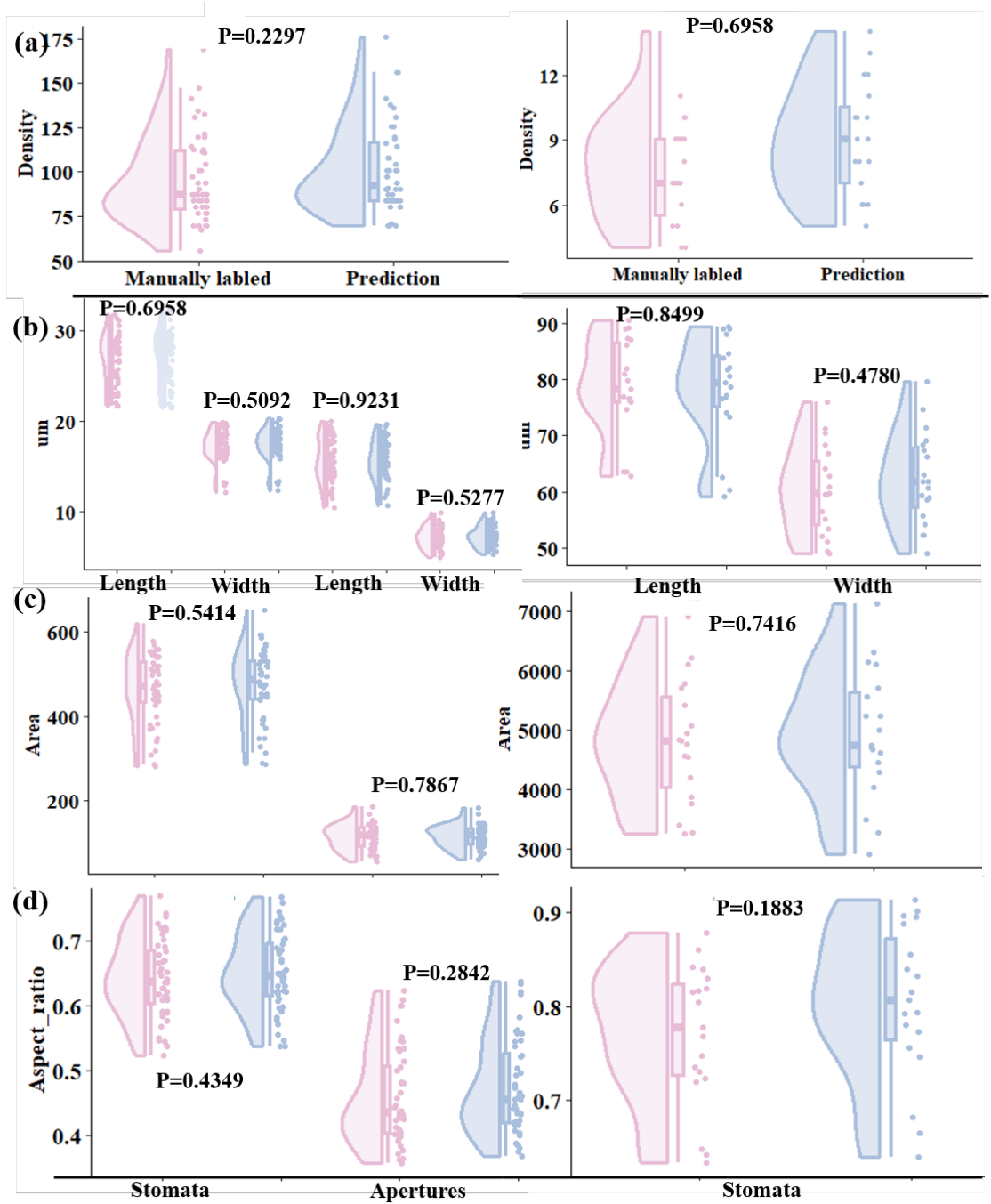

**Fig. 6** Statistic analysis. Violin plot-comparison of manually labeled data and StomaD$^2$ prediction data of peanut and maize. (a) density (b) width–length value (c) area (d) aspect ratio. The p-value represents the significance test between the two groups. A t-test is performed if the data follow a normal distribution; otherwise, the Wilcoxon rank-sum test is used. No significant differences are observed between manual annotations and prediction results.

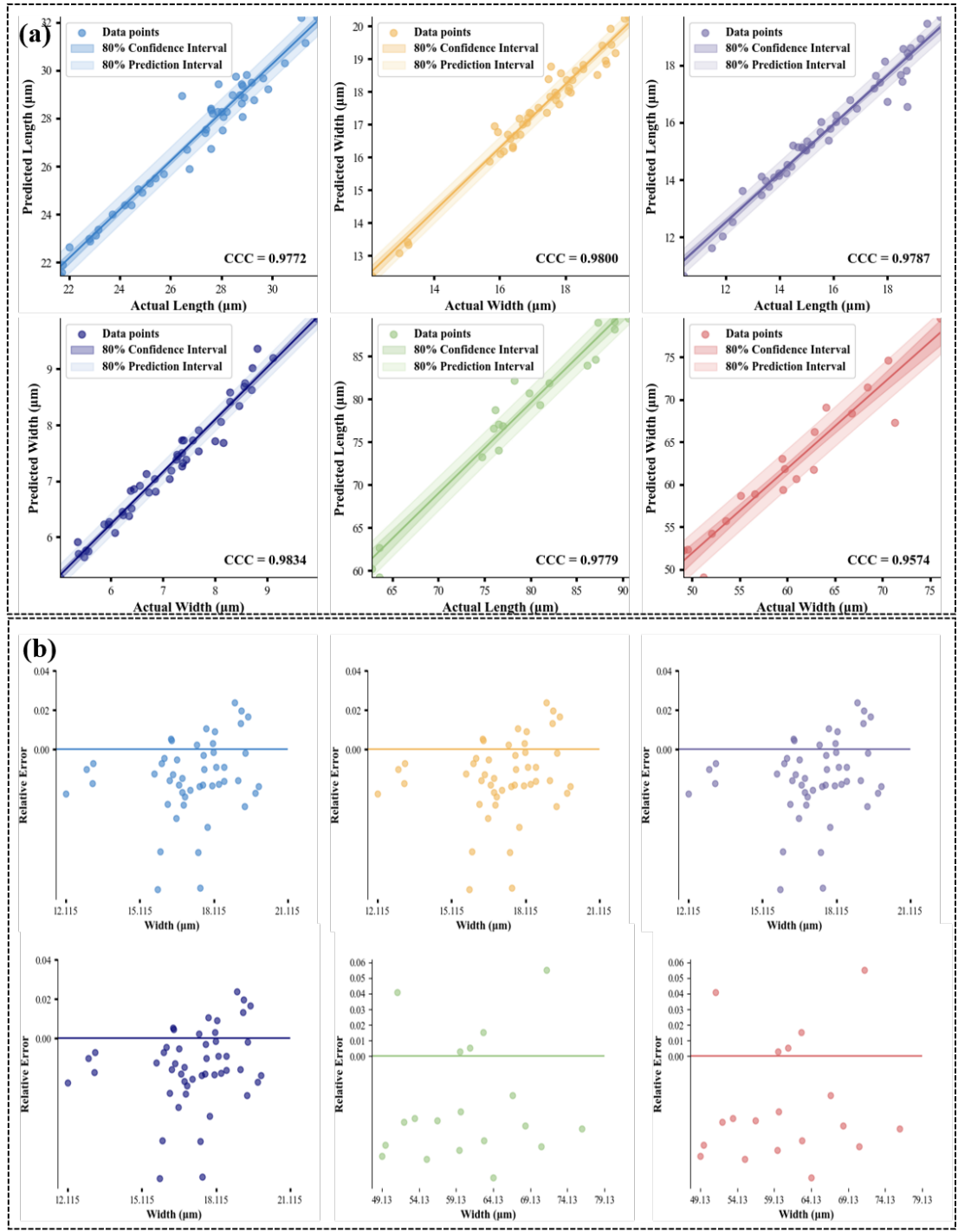

**Fig. 7** Comparison of manual and predicted values for peanut and maize. (a) Regression results for peanut stomatal length, stomatal width, aperture length, aperture width, and maize stomata length and width. (b) Relative errors for the corresponding traits.

**Table 7** Accuracy of length and width of stoma and aperture tests

| Dataset | Type | Avg. length accuracy(%) | Avg. width accuracy(%) | Length MSE | Width MSE | Length RMSE | Width RMSE |
|---|---|---|---|---|---|---|---|
| Peanut | Stomata | 0.9812 | 0.9842 | 0.3971 | 0.1745 | 0.6301 | 0.4177 |
|  | Aperture | 0.9694 | 0.9768 | 0.2918 | 0.0633 | 0.5402 | 0.2516 |
| Maize | Stomata | 0.9563 | 0.9788 | 3.9806 | 3.4806 | 1.9952 | 2.9121 |

**3.4 Evaluation of Generalization and Robustness**

The generalization ability of StomaD² was further validated by evaluating its performance on plant species not included in the training phase. Specifically, the model was trained solely on peanut data and then directly applied to images from soybean, crabapple, Arabidopsis, and fava beans—species it had not previously encountered. Fig. 8 presents qualitative detection results for a subset of these unseen species, showing that StomaD² consistently identifies stomata and apertures with high visual accuracy. Complementary to this, Table 8 provides quantitative detection metrics, demonstrating that the model maintains strong performance without the need for retraining or fine-tuning. These results highlight the robustness and adaptability of StomaD², confirming its generalization capability across both monocotyledonous and dicotyledonous plants, despite variations in stomatal morphology and image characteristics.

Table 8 Quantitative detection metrics in generalization experiments

|  | species | Stomata | Aperture | mAP | Stomata F1 | Aperture F1 |
|---|---|---|---|---|---|---|
| Dicotyledon | Broad bean | - | 0.978 | 0.978 | - | 0.955 |
|  | Peanut (destructive) | 0.980 | 0.926 | 0.953 | 0.945 | 0.911 |
|  | Soybean | 0.854 | 0.802 | 0.828 | 0.791 | 0.766 |
|  | Peanut (nondestructive) | 0.990 | 0.988 | 0.989 | 0.972 | 0.973 |
|  | Arabidopsis | 0.981 | 0.969 | 0.975 | 0.952 | 0.949 |
|  | Behonia | 0.984 | 0.982 | 0.983 | 0.96 | 0.958 |
| Monocotyledon | Maize (nondestructive) | 0.987 | - | 0.987 | 0.964 | - |
|  | Maize (destructive) | 0.977 | - | 0.977 | 0.955 | - |
|  | Wheat | 0.991 | - | 0.991 | 0.977 | - |
| Hybrid | 132 | 0.969 | - | 0.969 | 0.921 | - |

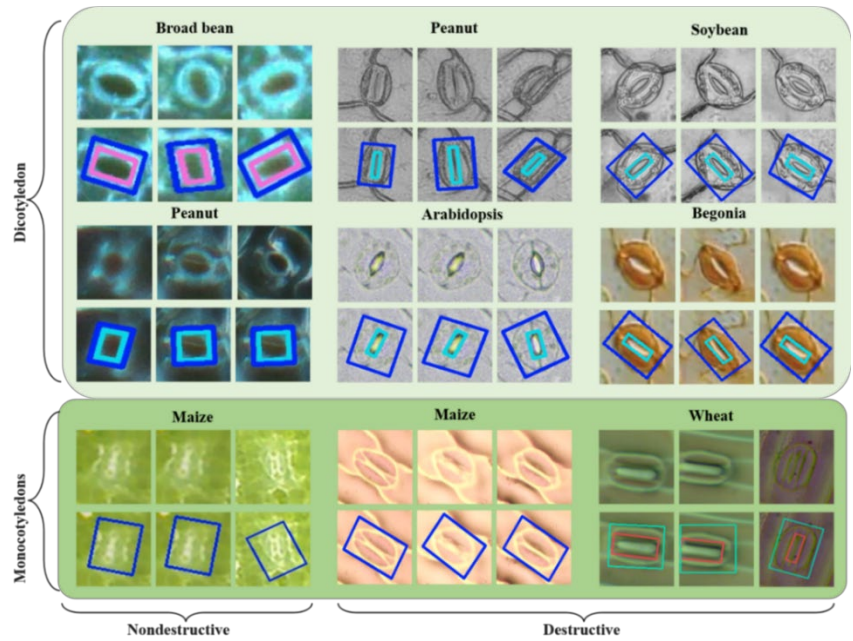

**Fig. 8** General Results. Stomatal and apertures detection results for different plants.

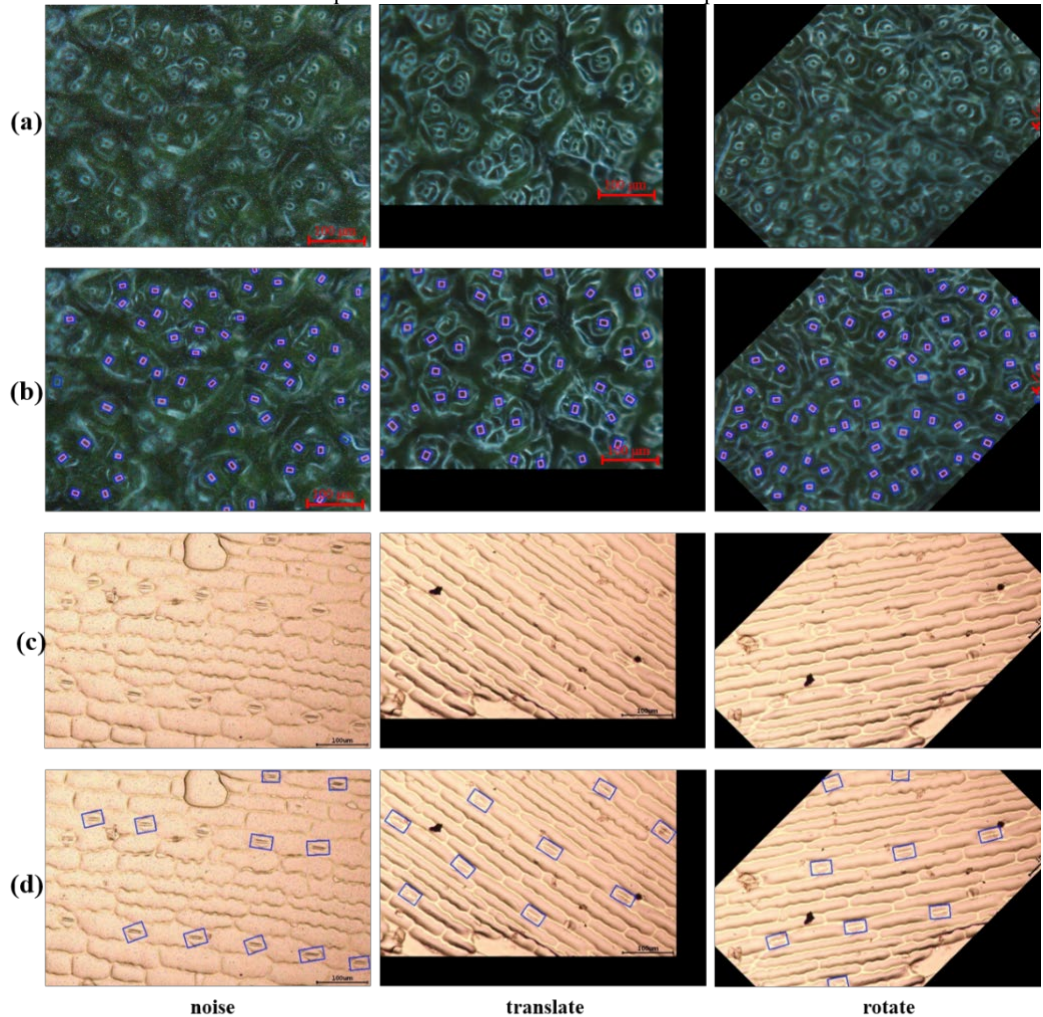

**Fig. 9** Robustness detection results. The first column shows data with added noise, the second column shows data with translation, and the third column shows data with rotation. (a) represents the processed data of peanuts, (b) represents the corresponding detection results of peanuts, (c) represents the processed data of maize, and (d) represents the corresponding detection results of maize.

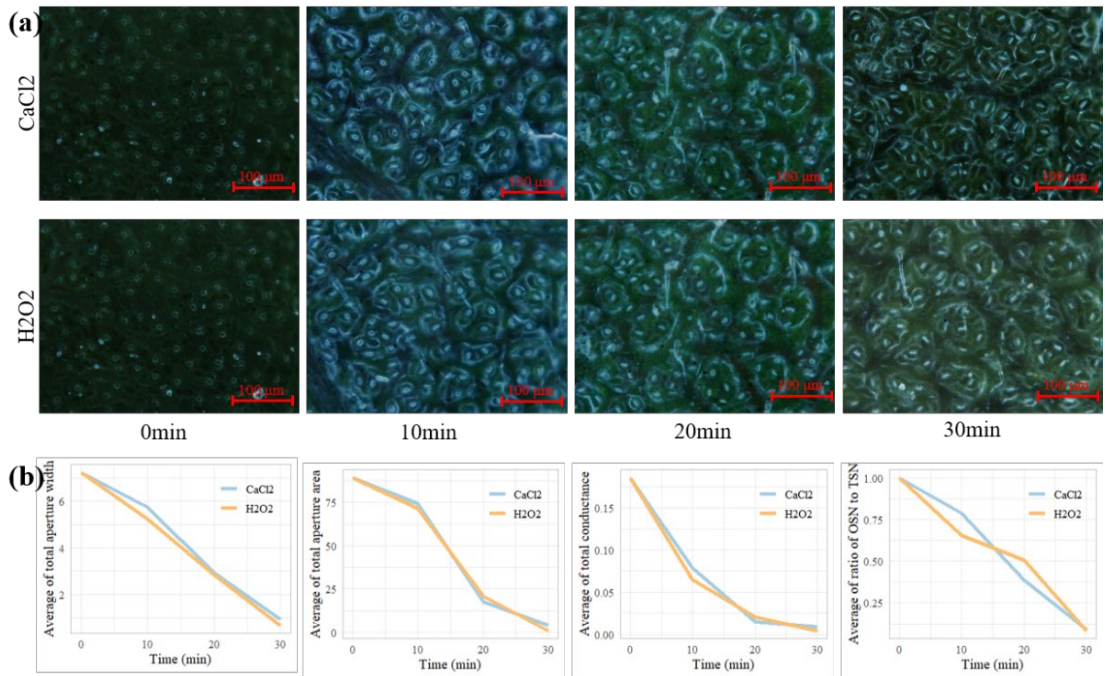

**Fig. 10** Time-course analysis of stomatal responses induced by exogenous $CaCl_2$ and $H_2O_2$ within 30 minutes. (a) Microscopic images of stomata treated with $CaCl_2$ and $H_2O_2$ at different time points. (b) Time-dependent curves showing changes in stomatal width, stomatal aperture area, stomatal conductance, and the ratio of opening stomata number (OSN) to total stomata number (TSN) under the two treatments.

**3.5 Evaluation of Stomatal Response Dynamics under Exogenous Treatments**

The dynamic monitoring of stomatal responses to exogenous $CaCl_2$ and $H_2O_2$ treatments in peanuts highlights the physiological relevance and application potential of StomaD² in real-time analysis of stomatal behavior (Fig. 10). Stomata play a pivotal role in regulating gas exchange and water loss, acting as gatekeepers in balancing photosynthetic efficiency and transpiration. Their rapid opening and closing in response to environmental cues are governed by complex signaling networks involving secondary messengers such as $Ca^{2+}$ and reactive oxygen species. Treatment with $CaCl_2$ and $H_2O_2$ induced progressive stomatal closure within a 30-minute period, consistent with their known functions as key signaling molecules in guard cell regulation. Calcium ions are widely recognized as universal second messengers that mediate stomatal closure through cytosolic calcium elevation, which activates downstream ion channels and leads to turgor loss in guard cells. Similarly, $H_2O_2$ is a central component of the reactive oxygen species signaling cascade.

**4 Discussion**

This study successfully developed and validated an integrated system, termed StomaD², for high-throughput and high-precision automated stomatal phenotyping. The system integrates advanced image restoration and detection algorithms into a user-friendly, standalone software package, significantly lowering the barrier to entry for researchers in plant science (Video A1 and Fig. 11). Experimental results demonstrate that StomaD² achieves high concordance with expert manual annotations. In a validation set comprising 68 microscopy images from peanut and maize, key stomatal traits such as density and aperture dimensions exhibited Concordance Correlation Coefficients (CCC) consistently exceeding 0.95, with relative errors generally below 0.06 (Fig. 6 and Fig. 7). These findings validate that StomaD² can achieve expert-level accuracy, establishing it as a reliable tool for automated analysis.

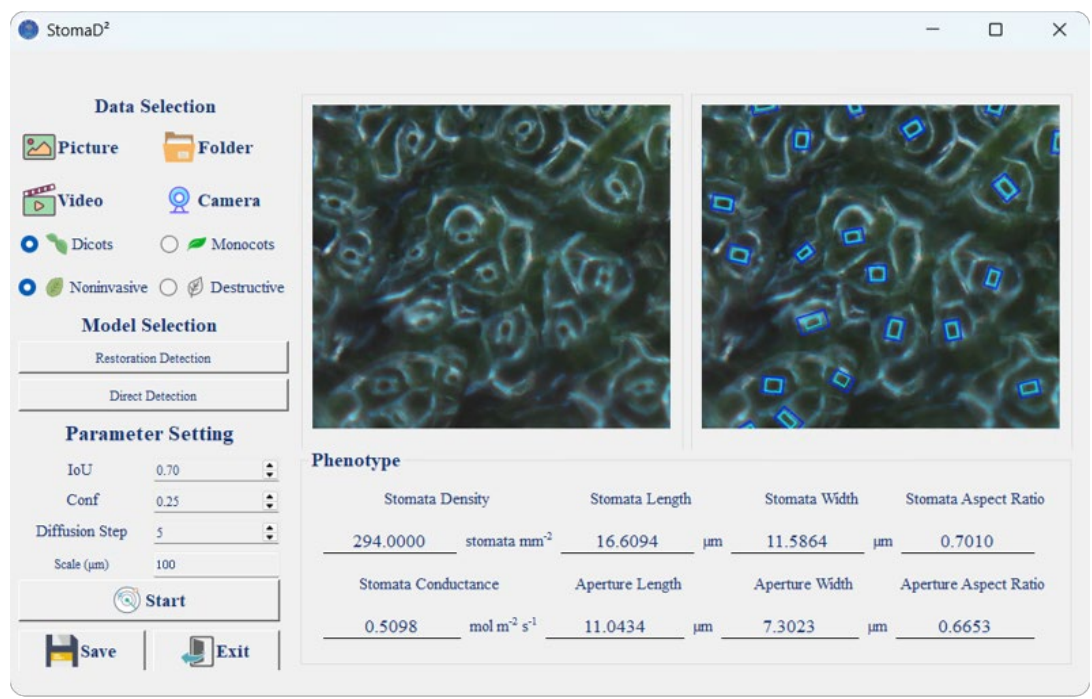

**Fig. 11** StomaD² system demonstration.

To further benchmark its performance against the SOTA, StomaD² was compared with multiple leading rotated object detection models. The results indicate that StomaD² exhibits superior performance, achieving a top-tier F1-score of 0.989 and a mAP of 0.973, outperforming prominent models such as Oriented Former (F1: 0.961) and YOLOv12 (F1: 0.982). Furthermore, on public maize and wheat datasets, the framework achieved accuracies of 0.994 and 0.992, respectively, consistently surpassing existing benchmarks like StomataHub. This quantitative evidence substantiates the superiority of the proposed network architecture for the specific challenges of detecting small, dense, and cluttered stomatal targets.

The exceptional performance of StomaD² can be attributed to its integrated "restoration-detection" framework and a highly specialized network architecture. The diffusion-based restoration module serves as a critical pre-processing step, effectively mitigating image degradations from blur, noise, and low-light conditions (Fig. 5). This provides a high-quality input for the subsequent detection network, thereby fundamentally reducing feature ambiguity, as evidenced by the heatmaps (Fig. 5 (c)). Furthermore, the detection network itself was specifically designed to address the unique characteristics of stomata. Unlike standard object detection frameworks, its horizontally spread, column-wise structure facilitates global feature interaction, capturing contextual information across the entire field of view. The multi-scale context-aware reweighting mechanism enhances cross-level semantic consistency, ensuring accurate detection across various scales, while the feature reassembly module strengthens intra-scale discrimination and suppresses complex background interference. This targeted design explains the model's significant performance gains over more generic architectures.

The broad applicability of this system, validated across more than 130 plant species, holds significant value for plant physiology, ecological research, and crop improvement. Stomatal traits are intrinsically linked to critical physiological processes including photosynthesis, transpiration, and stress responses. By enabling automated, high-throughput quantification of these features, StomaD² facilitates large-scale phenotypic comparisons and investigations into plant-environment interactions. The robustness of the framework was further confirmed through perturbation tests where images were subjected to noise, rotation, and translation ([Fig. 9](#)). The model maintained reliable detection performance across all perturbed datasets, demonstrating its resilience to common imaging artifacts encountered in practical applications. The ability to noninvasively track stomatal dynamics in intact leaves also provides a powerful tool for exploring the temporal patterns of plant stress adaptation in vivo.

Despite the outstanding performance of StomaD², this study has several limitations that also highlight promising avenues for future research. First, while the proposed restoration module effectively handles common image noise and low-pixel issues, its performance in extremely low-light or severely blurred conditions has room for improvement. To address this, future work will focus on exploring more advanced low-light image enhancement algorithms and developing lightweight designs capable of real-time restoration. Second, the system's real-time video detection function, particularly with high-resolution inputs, is computationally intensive, which may limit its direct deployment on low-power edge devices. Therefore, we plan to investigate model compression techniques, such as quantization and pruning, to develop a lightweight version of StomaD² suitable for a wide range of edge devices. Finally, for certain species with complex three-dimensional surface structures, such as dense epidermal trichomes or sunken stomata, the accuracy of the current 2D-based detection may be affected. To overcome this challenge, our future research will explore the fusion of multispectral or 3D microscopy data to more accurately differentiate and quantify stomata on these complex surfaces, further enhancing the model's generalizability and precision.

## 5 Conclusion

This paper presented StomaD², a diffusion-based restoration-detection framework that significantly improves the performance of stomatal detection in challenging images captured by handheld microscopes. The framework enables fast, high-throughput detection and phenotypic analysis of live stomata and is packaged as a user-friendly, standalone executable to ensure broad accessibility. The robustness and accuracy of StomaD² were validated against expert manual annotations, demonstrating high consistency with CCC exceeding 0.95 and relative errors below 0.06 for key stomatal traits. Furthermore, when benchmarked against SOTA models, StomaD² achieved a superior F1-score of 0.989, confirming its advanced performance. On public maize and wheat datasets, the model also demonstrated excellent generalizability. By liberating researchers from labor-intensive and low-throughput measurement tasks, StomaD² offers a powerful and reliable tool that can accelerate physiological and ecological studies, holding significant value for plant-environment interaction research and crop improvement.

## References

[1] Hetherington, A.M., Woodward, F.I., 2003. The role of stomata in sensing and driving environmental change. Nature 424, 901–908. https://doi.org/10.1038/nature01843.

[2] Casado-García, A., del-Canto, A., Sanz-Saez, A., Pérez-López, U., Bilbao-Kareaga, A., Fritschi, F.B., Miranda-Apodaca, J., Muñoz-Rueda, A., Sillero-Martínez, A., Yoldi-Achalandabaso, A., Lacuesta, M., Heras, J., 2020. LabelStoma: A tool for stomata detection based on the YOLO algorithm. Computers and Electronics in Agriculture 178, 105751.

https://doi.org/10.1016/j.compag.2020.105751.

[3] Medeiros, D.B., Martins, S.C.V., Cavalcanti, J.H.F., Daloso, D.M., Martinoia, E., Nunes-Nesi, A., DaMatta, F.M., Fernie, A.R., Araújo, W.L., 2016. Enhanced photosynthesis and growth in atquac1 knockout mutants are due to altered organic acid accumulation and an increase in both stomatal and mesophyll conductance. Plant Physiology 170, 86–101. https://doi.org/10.1104/pp.15.01375.

[4] Faralli, M., Matthews, J., Lawson, T., 2019. Exploiting natural variation and genetic manipulation of stomatal conductance for crop improvement. Current Opinion in Plant Biology 49, 1–7. https://doi.org/10.1016/j.pbi.2019.03.004.

[5] Tuberosa, R., 2012. Phenotyping for drought tolerance of crops in the genomics era. Frontiers in Physiology 3, 347. https://doi.org/10.3389/fphys.2012.00347.

[6] Bertolino, L.T., Cabañero, S., Gray, J.E., 2019. Impact of stomatal density and morphology on water-use efficiency in a changing world. Frontiers in Plant Science 10, 225. https://doi.org/10.3389/fpls.2019.00225.

[7] Kim, T.H., Böhmer, M., Hu, H., Nishimura, N., Schroeder, J.I., 2010. Guard cell signal transduction network: advances in understanding abscisic acid, $Co_2$, and $Ca^{2+}$ signaling. Annual Review of Plant Biology 61, 561–591. https://doi.org/10.1146/annurev-arplant-042809-112226.

[8] Shimazaki, K., Doi, M., Assmann, S.M., Kinoshita, T., 2007. Light regulation of stomatal movement. Annual Review of Plant Biology 58, 219–247. https://doi.org/10.1146/annurev.arplant.58.032806.103831.

[9] Xu, B., Zhang, J., Tang, Z., Zhang, Y., Xu, L., Lu, H., Han, Z., Hu, W., 2025. Nighttime environment enables robust field-based high-throughput plant phenotyping: A system platform and a case study on rice. Computers and Electronics in Agriculture 235, 110337. https://doi.org/10.1016/j.compag.2025.110337.

[10] Igboabuchi, N.A., Ilodibia, C.V., 2017. A study on the anatomy of Zanthoxylum macrophylla (Rutaceae). Asian Journal of Biology 5, 1–5. https://doi.org/10.9734/AJOB/2017/36184.

[11] Bourdais, G., McLachlan, D.H., Rickett, L.M., Zhou, J., Siwoszek, A., Häweker, H., Hartley, M., Kuhn, H., Morris, R.J., MacLean, D., Hetherington, A.M., Zipfel, C., 2019. The use of quantitative imaging to investigate regulators of membrane trafficking in Arabidopsis stomatal closure. Traffic 20, 168–180. https://doi.org/10.1111/tra.12634.

[12] Millstead, L., Jayakody, H., Patel, H., Kaura, V., Petrie, P.R., Tomasetig, F., Whitty, M., 2020. Accelerating automated stomata analysis through simplified sample collection and imaging techniques. Frontiers in Plant Science 11, 580389. https://doi.org/10.3389/fpls.2020.580389.

[13] Sun, Z.Z., Song, Y.L., Li, Q., Cai, J., Wang, X., Zhou, Q., Huang, M., Jiang, D., 2021. An integrated method for tracking and monitoring stomata dynamics from microscope videos. Plant Phenomics 2021, 9892647. https://doi.org/10.34133/2021/9892647.

[14] Kwong, Q.B., Wong, Y.C., Lee, P.L., Sahaini, M.S., Kon, Y.T., Kulaveerasingam, H., Appleton, D.R., 2021. Automated stomata detection in oil palm with convolutional neural network. Scientific Reports 11, 15210. https://doi.org/10.1038/s41598-021-94520-x.

[15] Yang, X.H., Wang, Y.T., Wu, M.H., Li, F., Zhou, C.L., Yang, L.J., Zheng, C., Li, Y., Li, Z., Guo, S.Y., Song, C.P., 2024a. SLPA-Net: a real-time recognition network for intelligent stomata

localization and phenotypic analysis. IEEE/ACM Transactions on Computational Biology and Bioinformatics 21, 372–382. https://doi.org/10.1109/TCBB.2023.3242279.

[16] Solimani, F., Cardellicchio, A., Dimauro, G., Petrozza, A., Summerer, S., Cellini, F., Renò, V., 2024. Optimizing tomato plant phenotyping detection: Boosting YOLOv8 architecture to tackle data complexity. Computers and Electronics in Agriculture 218, 108728. https://doi.org/10.1016/j.compag.2024.108728.

[17] Fetter, K.C., Eberhardt, S., Barclay, R.S., Wing, S., Keller, S.R., 2019. StomataCounter: a neural network for automatic stomata identification and counting. New Phytologist 223, 1671–1681. https://doi.org/10.1111/nph.15892.

[18] Zhang, F., Wang, B., Lu, F.H., Zhang, X.H., 2023. Rotating stomata measurement based on anchor-free object detection and stomata conductance calculation. Plant Phenomics 5, 0106. https://doi.org/10.34133/plantphenomics.0106.

[19] Gu, Z., Ma, X., Guan, H., Jiang, Q., Deng, H., Wen, B., Zhu, T., Wu, X., 2024. Tomato fruit detection and phenotype calculation method based on the improved RTDETR model. Computers and Electronics in Agriculture 227, 109524. https://doi.org/10.1016/j.compag.2024.109524.

[20] Yang, X.H., Wang, J.H., Li, F., Zhou, C.L., Zheng, C., Yang, L.J., Li, Z., Li, Y., Guo, S.Y., Song, C.P., Li, G., 2024. RotatedStomataNet: a deep rotated object detection network for directional stomata phenotype analysis. Plant Cell Reports 43, 108. https://doi.org/10.1007/s00299-024-03173-z.

[21] Liang, X.Y., Xu, X.C., Wang, Z.W., He, L., Zhang, K.Q., Liang, B., Ye, J.L., Shi, J.W., Wu, X., Dai, M.Q., Zhou, J.J., Wang, Z.Y., Wang, X.M., Zhang, J.Y., Wu, J., Lin, Y.J., 2022. StomataScorer: a portable and high-throughput leaf stomata trait scorer combined with deep learning and an improved CV model. Plant Biotechnology Journal 20, 577–591. https://doi.org/10.1111/pbi.13745.

[22] Song, W.L., Li, J.Y., Li, K.X., Chen, J.X., Huang, J.P., 2020. An automatic method for stomatal pore detection and measurement in microscope images of plant leaf based on a convolutional neural network model. Forests 11, 954. https://doi.org/10.3390/f11090954.

[23] Lin, X.Q., He, J.W., Chen, Z.Y., Lyu, Z.Y., Dai, B., Yu, F.H., Qiao, Y., Ouyang, W.L., Dong, C., 2022. Diffbir: towards blind image restoration with generative diffusion prior. In: Proceedings of the European Conference on Computer Vision, 430–448. https://doi.org/10.1007/978-3-031-19787-1_25.

[24] Deng, J., Dong, W., Socher, R., Li, L.J., Li, K., Li, F.F., 2009. ImageNet: a large-scale hierarchical image database. In: Proceedings of the IEEE Conference on Computer Vision and Pattern Recognition, 248–255. https://doi.org/10.1109/CVPR.2009.5206848.

[25] Liang, J.Y., Cao, J.Z., Sun, G.L., Zhang, K., Van Gool, L., Timofte, R., 2021. SwinIR: image restoration using swin transformer. In: Proceedings of the IEEE/CVF International Conference on Computer Vision, 1833–1844. https://doi.org/10.1109/ICCV48922.2021.00185.

[26] Rombach, R., Blattmann, A., Lorenz, D., Esser, P., Ommer, B., 2022. High-resolution image synthesis with latent diffusion models. In: Proceedings of the IEEE/CVF Conference on Computer Vision and Pattern Recognition, 10684–10695. https://doi.org/10.1109/CVPR52688.2022.01042.

[27] Diederik, P.K., Max, W., 2013. Auto-encoding variational bayes. *International Conference on Learning Representations*.

[28] Wang, C.Y., Liao, H.Y.M., Wu, Y.H., Chen, P.Y., Hsieh, J.W., Yeh, I.H., 2020. CSPNet: a new backbone that can enhance learning capability of CNN. In: Proceedings of the IEEE/CVF Conference on Computer Vision and Pattern Recognition Workshops, 1571–1580. https://doi.org/10.1109/CVPRW50498.2020.00203.

[29] Muhammad, M.B., Yeasin, M., 2020. Eigen-CAM: class activation map using principal components. In: 2020 International Joint Conference on Neural Networks, 1–7. https://doi.org/10.1109/IJCNN48605.2020.9207424.

[30] Schuhmann, C., Beaumont, R., Vencu, R., Gordon, C., Wightman, R., Cherti, M., Coombes, T., Katta, A., Mullis, C., Wortsman, M., Schramowski, P., Kundurthy, S., Crowson, K., Schmidt, L., Kaczmarczyk, R., Jitsev, J., 2022. LAION-5B: an open large-scale dataset for training next generation image-text models. Advances in Neural Information Processing Systems 35, 25278–25294.

[31] Sack, L., Buckley, T.N., 2016. The developmental basis of stomatal density and flux. Plant Physiology 171, 2358–2363. https://doi.org/10.1104/pp.16.00773.

[32] Murray, M., Soh, W.K., Yiotis, C., Spicer, R., Lawson, T., McElwain, J.C., 2020. Consistent relationship between field-measured stomatal conductance and theoretical maximum stomatal conductance in $C_3$ woody angiosperms in four major biomes. International Journal of Plant Sciences 181, 142–154. https://doi.org/10.1086/706222.

[33] Yang, N., Huang, Z., He, Y., Xiao, W., Yu, H., Qian, L., Xu, Y., Tao, Y., Lyu, P., Lyu, X., Feng, X., 2024. Detection of color phenotype in strawberry germplasm resources based on field robot and semantic segmentation. Computers and Electronics in Agriculture 226, 109464. https://doi.org/10.1016/j.compag.2024.109464.

[34] Franks, P.J., Beerling, D.J., 2009. Maximum leaf conductance driven by $CO_2$ effects on stomatal size and density over geologic time. Proceedings of the National Academy of Sciences 106, 10343–10347. https://doi.org/10.1073/pnas.0904201106.

[35] Koho, S., Fazeli, E., Eriksson, J.E., Hänninen, P.E., 2016. Image quality ranking method for microscopy. Scientific Reports 6, 28962. https://doi.org/10.1038/srep28962.

[36] Ren, S., He, K., Girshick, R., & Sun, J. (2016). Faster R-CNN: Towards real-time object detection with region proposal networks. *IEEE transactions on pattern analysis and machine intelligence*, *39*(6), 1137-1149. https://doi.org/10.1109/TPAMI.2016.2577031.

[37] Xie, X.X., Cheng, G., Wang, J.B., Yao, X.W., Han, J.W., 2021. Oriented R-CNN for object detection. In: Proceedings of the IEEE/CVF International Conference on Computer Vision, 3520–3529. https://doi.org/10.1109/ICCV48922.2021.00352.

[38] Yang, X., Yan, J.C., Feng, Z.M., He, T., 2021. R3Det: refined single-stage detector with feature refinement for rotating object. Proceedings of the AAAI Conference on Artificial Intelligence 35, 3163–3171. https://doi.org/10.1609/aaai.v35i4.16426.

[39] Zhao, J.Q., Ding, Z.Y., Zhou, Y., Zhu, H.C., Du, W.L., Yao, R., Saddik, A.E., 2024. OrientedFormer: an end-to-end transformer-based oriented object detector in remote sensing

images. IEEE Transactions on Geoscience and Remote Sensing 62, 5603816. https://doi.org/10.1109/TGRS.2024.3353721.

[40] Varghese, R., Sambath, M., 2024. YOLOv8: a novel object detection algorithm with enhanced performance and robustness. In: 2024 International Conference on Advances in Data Engineering and Intelligent Computing Systems, 1–6. https://doi.org/10.1109/ADICS60785.2024.10492868/

[41] Tian, Y.J., Ye, Q.X., Doermann, D. 2025. YOLOv12: Attention-centric real-time object detectors. *arXiv preprint arXiv:2502.12524*.

[42] Qin, M.E., Song, Y., Zhao, Q., Yang, X., Che, Y. and Yang, X., 2026. A3-FPN: Asymptotic Content-Aware Pyramid Attention Network for Dense Visual Prediction. *arXiv preprint arXiv:2604.10210*.

[43] Wang, J.X., Renninger, H.J., Ma, Q., Jin, S.C., 2024. Measuring stomatal and guard cell metrics for plant physiology and growth using StoManager1. Plant Physiology 195, 378–394. https://doi.org/10.1093/plphys/kiad688

[44] Gibbs, J.A., McAusland, L., Robles-Zazueta, C.A., Murchie, E.H., Burgess, A.J., 2021. A deep learning method for fully automatic stomatal morphometry and maximal conductance estimation. Frontiers in Plant Science 12, 780180. https://doi.org/10.3389/fpls.2021.780180.

[45] Gibbs, J.A., Gibbs, A.J., 2025. Integrating phenotyping and modelling approaches—StomaGAN: improving image-based analysis of stomata through generative adversarial networks. in silico Plants 7, diaf002. https://doi.org/10.1093/insilicoplants/diaf002.

[46] Liang, X.Y., Xu, X.C., Wang, Z.W., He, L., Zhang, K.Q., Liang, B., Ye, J.L., Shi, J.W., Wu, X., Dai, M.Q., Zhou, J.J., Wang, Z.Y., Wang, X.M., Zhang, J.Y., Wu, J., Lin, Y.J., 2022. StomataScorer: a portable and high-throughput leaf stomata trait scorer combined with deep learning and an improved CV model. *Plant Biotechnology Journal* 20, 577–591. https://doi.org/10.1111/pbi.13745.

[47] Wang, L., Guo, Y., Zhang, Z., Qin, M. E., Li, Z., Sun, X., & Yang, X. (2025). OS-MSWGBM: Intelligent Analysis of Organic Synthesis Based on Multiscale Subtraction Weighted Network and LightGBM. *MATCH-COMMUNICATIONS IN MATHEMATICAL AND IN COMPUTER CHEMISTRY*, *93*(1). https://doi.org/10.46793/match.93-1.005W.

[48] Guo, Y., Peng, L., Li, Z., Qin, M. E., Jiao, X., Chai, Y., & Yang, X. (2024). OCS-TGBM: Intelligent Analysis of Organic Chemical Synthesis Based on Topological Data Analysis and LightGBM. *MATCH-COMMUNICATIONS IN MATHEMATICAL AND IN COMPUTER CHEMISTRY*, *91*(3), 557-592. https://doi.org/10.46793/match.91-3.557G.

## Supporting Information

**Evaluation Metrics**

The accuracy of the model in detecting stoma or stoma openings is the main metric for evaluating the model's performance. Here, common metrics are used to evaluate the model's performance, including Precision, Recall, F1 score, Average Precision (AP), and mean Average Precision (mAP).

$$Precision = \frac{N_{TP}}{N_{TP} + N_{FP}},$$

$$Recall = \frac{N_{TP}}{N_{TP} + N_{FN}},$$

$$F1score = 2 \times \frac{Precision \times Recall}{Precision + Recall},$$

$$AP = \int_0^1 P(i)dR(i),$$

$$mAP = \frac{1}{n}\sum_{i=1}^{n} AP_i,$$

where $N_{TP}$ represents the number of correctly detected stoma or stoma openings, $N_{FP}$ is the number of false positives (regions incorrectly identified as stoma or stoma openings), $N_{TN}$ refers to the number of correctly detected background regions, and $N_{FN}$ represents the number of false negatives (stoma or stoma openings incorrectly identified as background), $n$ is the count and $i$ denotes the current object.

To evaluate the consistency between the phenotypic data predicted by StomaD² and those measured by researchers, we use the consistency correlation coefficient (CCC), which ranges from −1 to 1. As stomata and their apertures typically appear elliptical, the long axis represents the stomatal length, and the short axis represents the width. The detection box is the outer rectangle enclosing the stomata or apertures, with its length and width corresponding to the stomatal dimensions. In addition to metrics for detection and counting accuracy, we introduce several evaluation metrics, including length accuracy (Acc. length), width accuracy (Acc. width), average length accuracy (Avg. length accuracy), average width accuracy (Avg. width accuracy), mean square error (MSE), root mean square error (RMSE), and the area (Area) of the stomata or apertures within the detection or ground-truth boxes. The formulas for these calculations are provided below.

$$CCC = \frac{2\sigma_{(g_i,d_i)}}{(\mu_{gi},\mu_{di})^2 + \sigma_{gi}^2 + \sigma_{di}^2},$$

$$Acc_{width} = 1 - \frac{g_{wi} - d_{wi}}{g_{wi}},$$

$$Acc_{length} = 1 - \frac{g_{li} - d_{li}}{g_{li}},$$

$$Acc.widthaccuracy = 1 - \frac{1}{n}\sum_{wi=1}^{n}\frac{|g_{wi} - d_{wi}|}{g_{wi}},$$

$$Acc.lengthaccuracy = 1 - \frac{1}{n}\sum_{wi=1}^{n}\frac{|g_{li} - d_{li}|}{g_{li}},$$

$$MSE = \frac{1}{n}\sum_{wi=1}^{n}(g_i - d_i)^2,$$

$$RMSE = 1 - \frac{\sqrt{\sum_{wi=1}^{n}(g_i - d_i)^2}}{n},$$

$$Area_{i=}w_i \times l_i,$$

where, $g_i$ and $d_i$ represent the manually labelled data and the length or width values of the predicted stomata or apertures of the StomaD², the means of $g_i$ and di are expressed as $\mu_{gi}$ and $\mu_{di}$, the standard deviations are expressed as $\sigma_{g_i}$ and $\sigma_{d_i}$, and the covariance between them is expressed as $\sigma_{(g_i,d_i)}$, $l_i$ is the length value, $w_i$ represents the width.

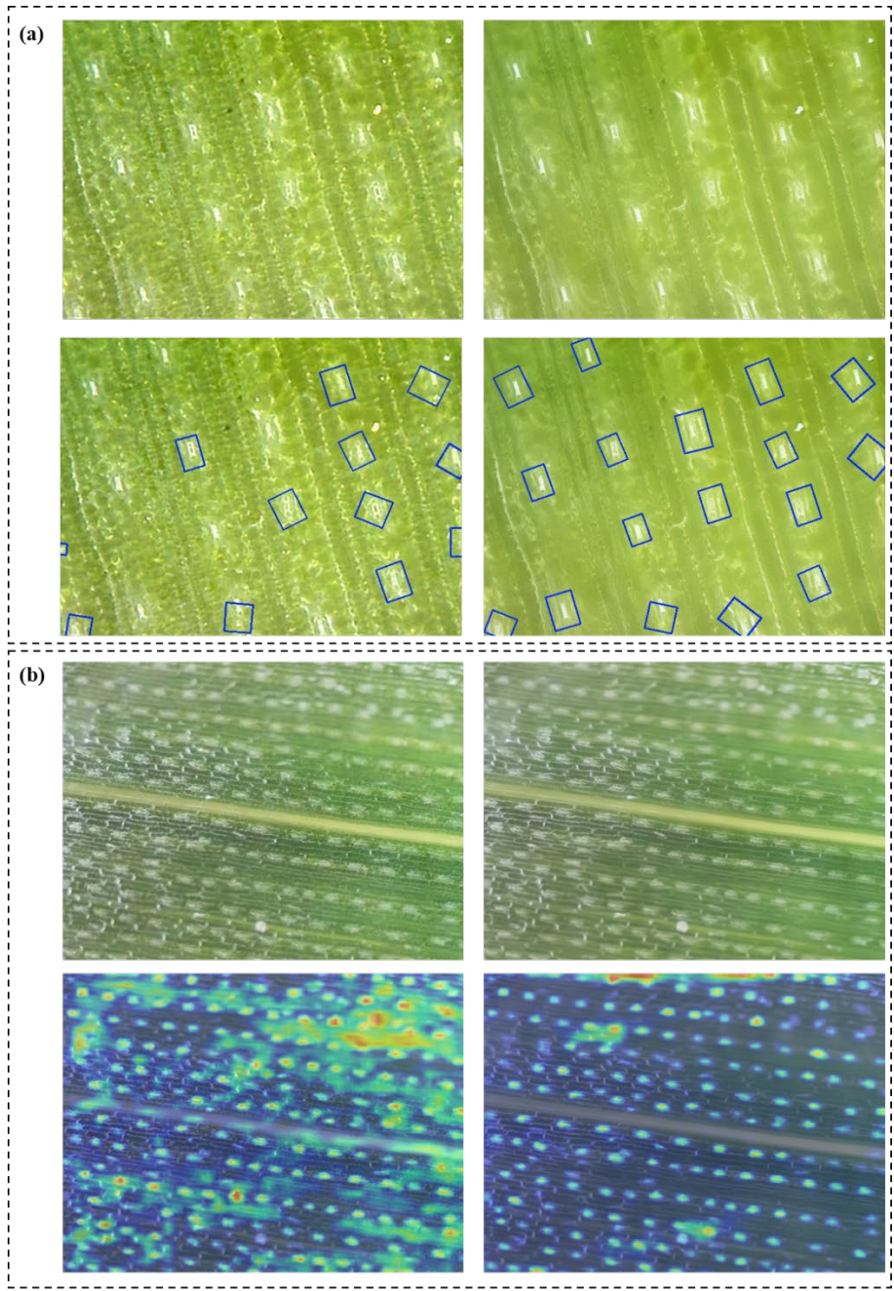

**Fig. A 1** Restoration efficacy. (a) compares noninvasive stomata detection on maize with post-restoration detection (b) shows the heatmap before restoration and the heatmap after restoration.

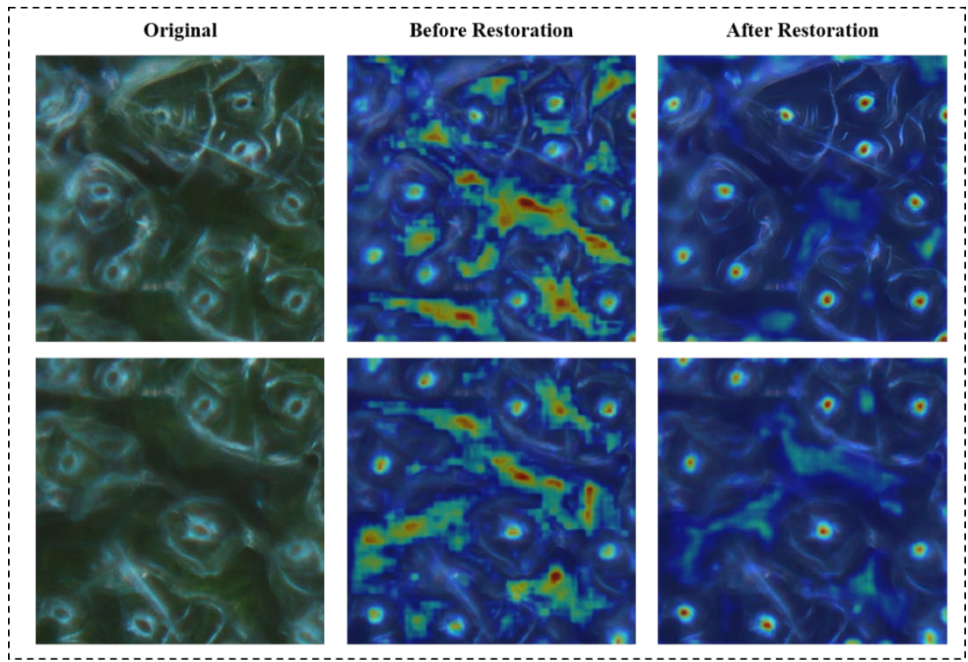

**Fig. A 2** Heatmaps of peanut before and after restoration.

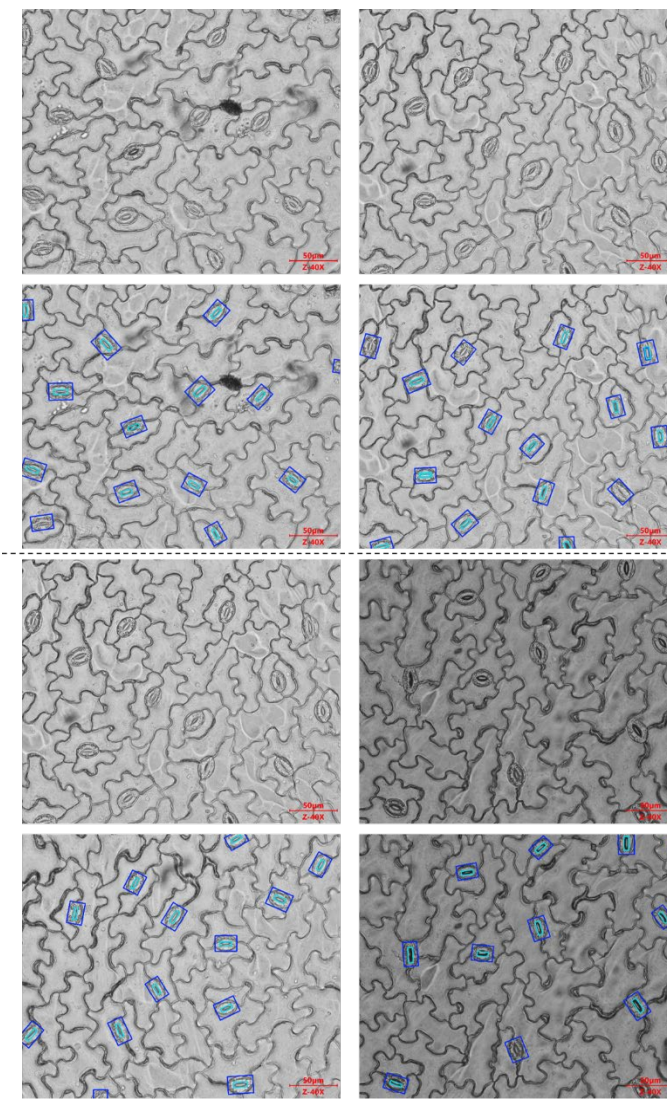

**Fig. A 3** Detection results of soybean stomata.

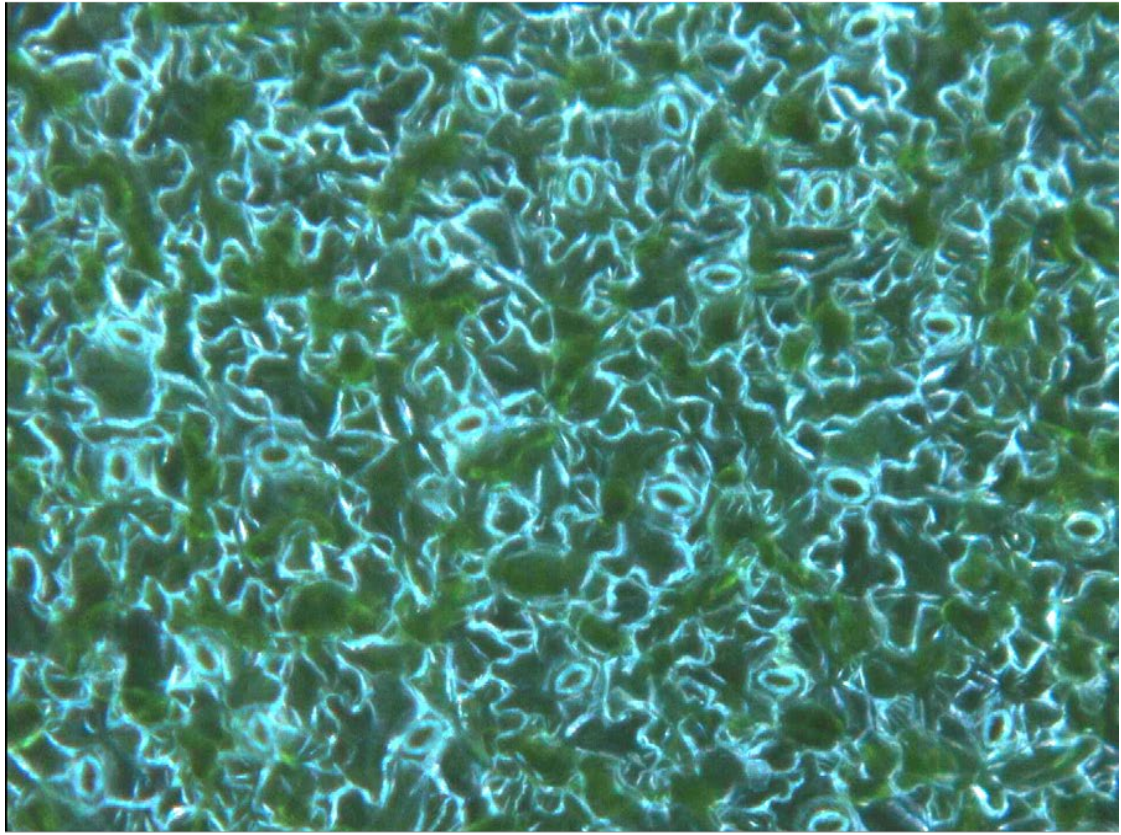
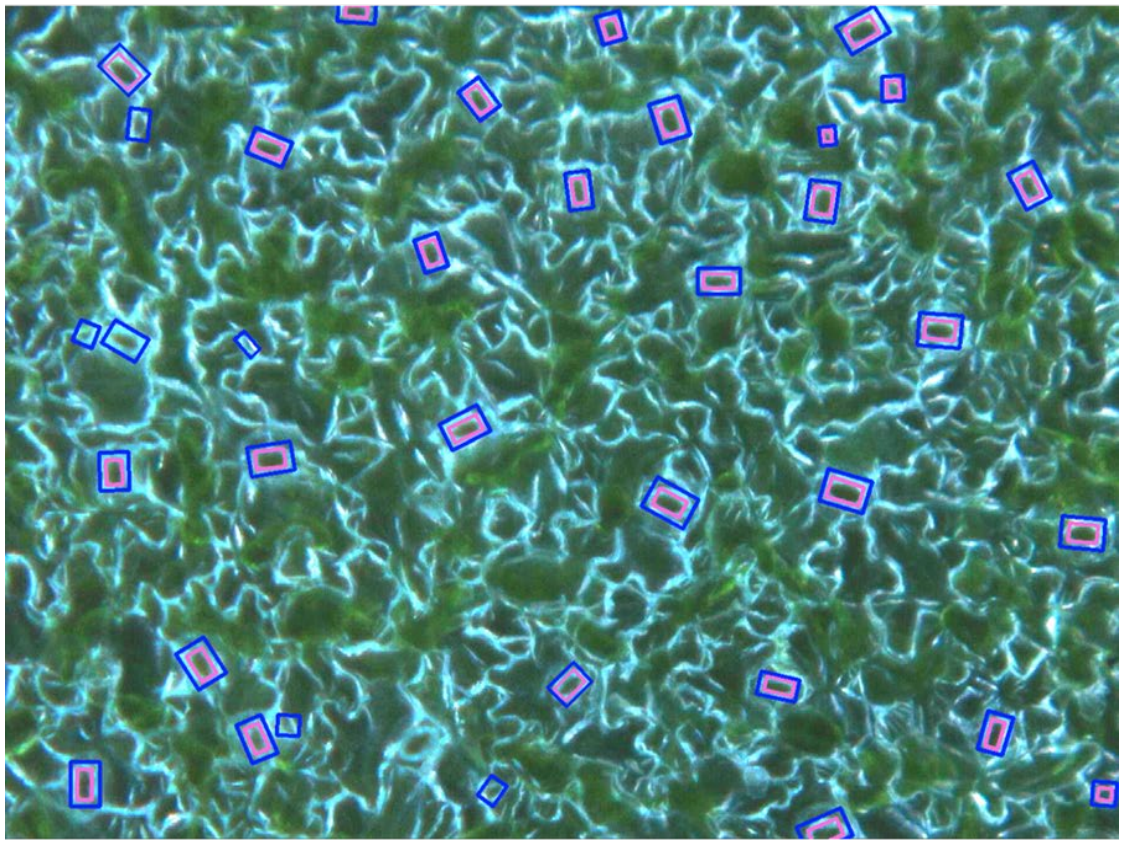

**Fig. A 4** Detection results of broad bean stomata.

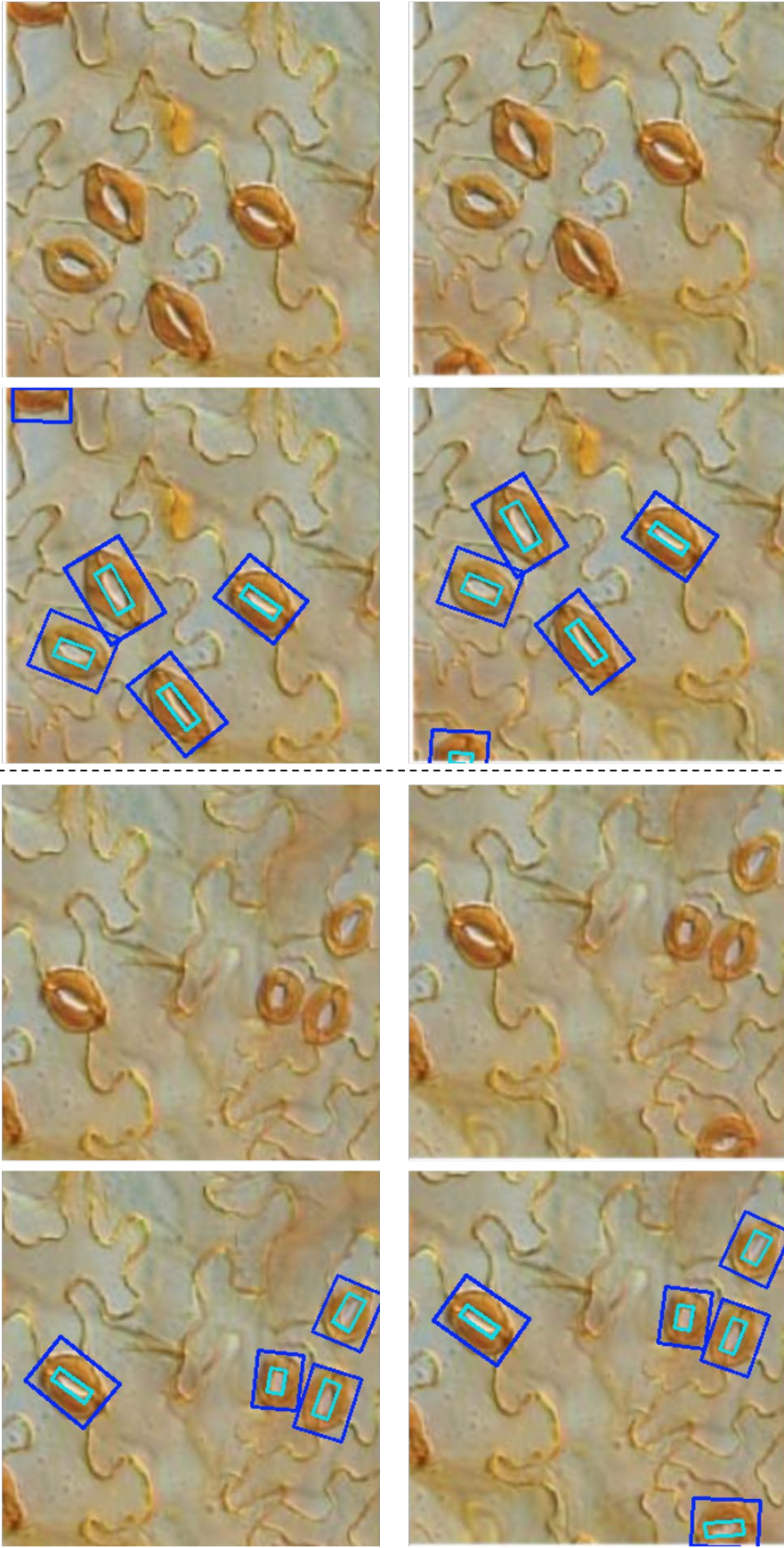

**Fig. A 5** Detection results of begonia stomata.

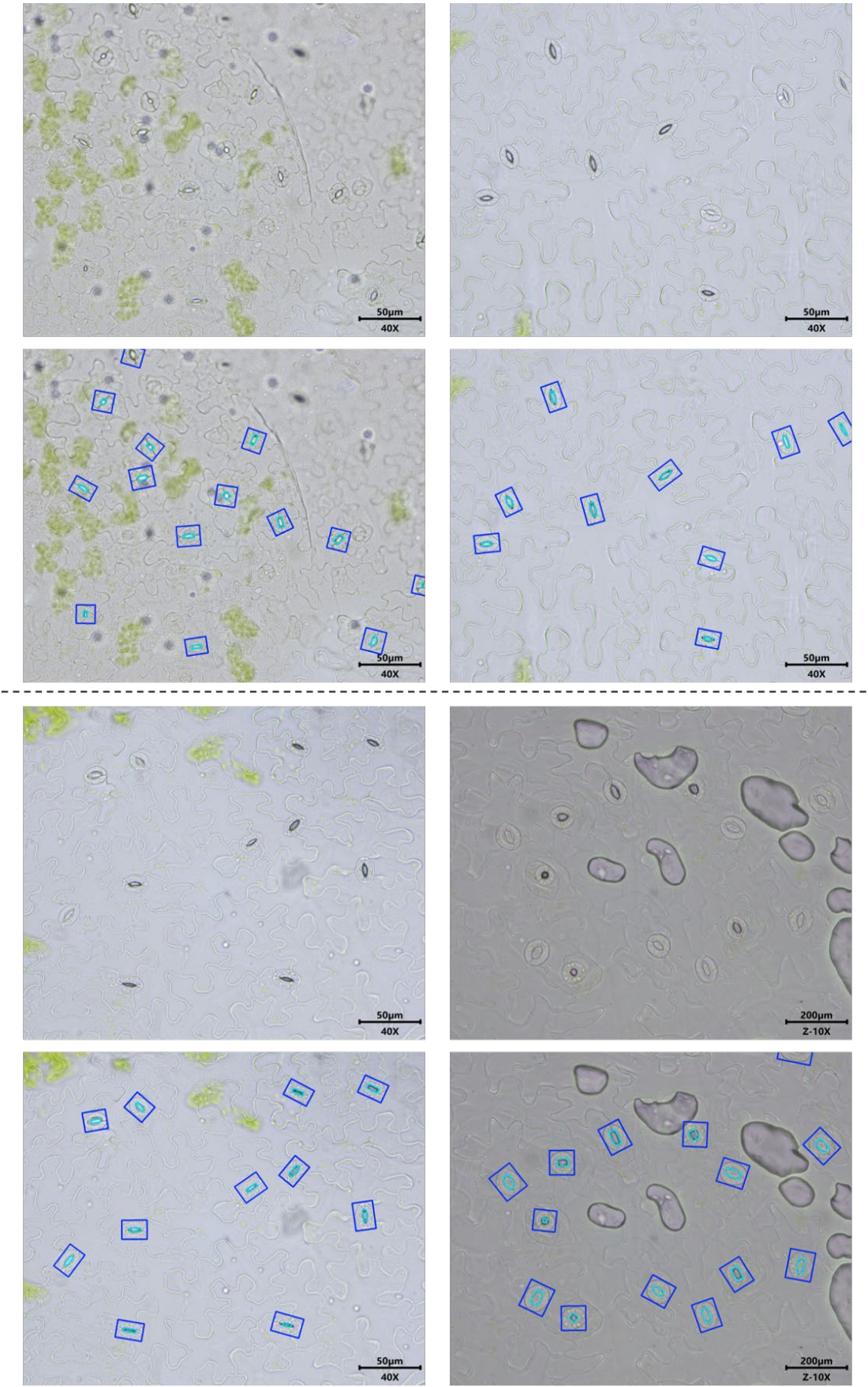

**Fig. A 6** Detection results of Arabidopsis stomata.

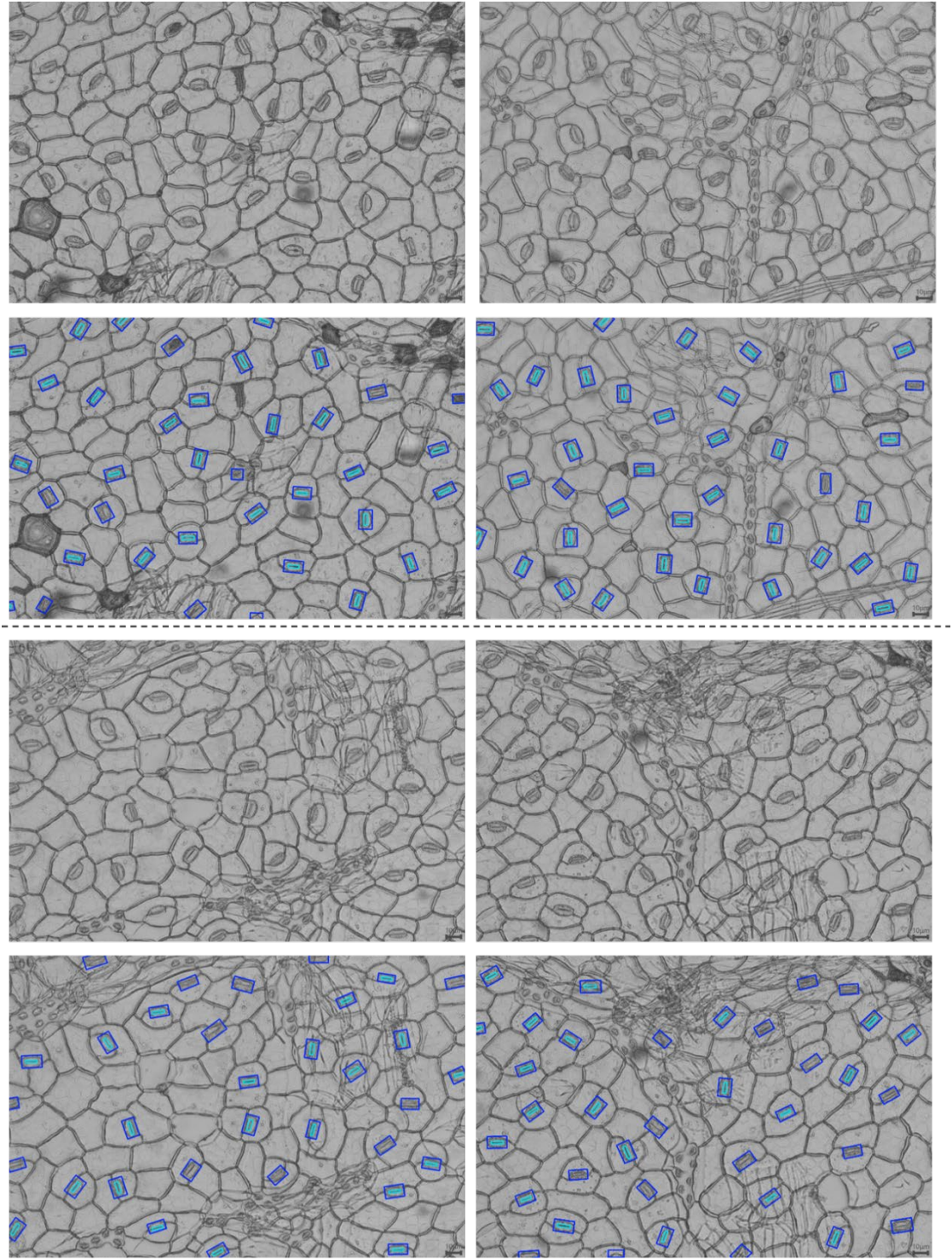

**Fig. A 7** Detection results of peanut stomata.

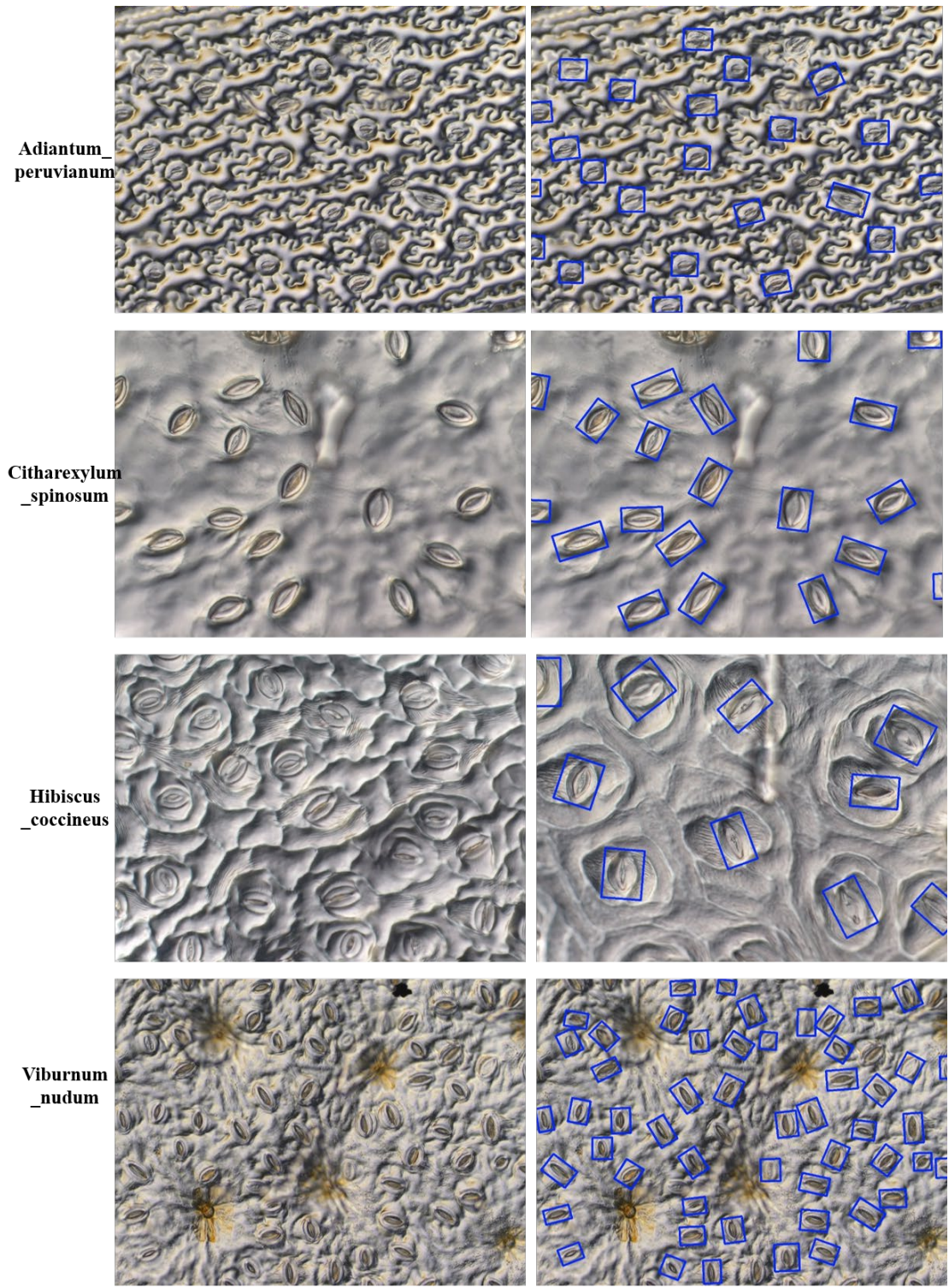

**Fig. A 8** Example results of multi-species generalization.

**Table A 1** Detection Results of StomaD$^2$

| stomata average height (μm) | stomata average width (μm) | stomata aspect ratio | aperture average height (μm) | aperture average width (μm) | aperture aspect ratio | stomata density (stomata * mm-2) | conductance (mol m-2 s-1) |
|---|---|---|---|---|---|---|---|
| 20.00 | 24.60 | 0.82 | 11.74 | 15.92 | 0.74 | 141.56 | 0.42 |
| 20.03 | 24.59 | 0.82 | 11.83 | 16.10 | 0.74 | 147.00 | 0.44 |
| 20.03 | 24.46 | 0.82 | 11.85 | 16.06 | 0.74 | 145.19 | 0.44 |
| 19.86 | 24.55 | 0.81 | 11.72 | 16.06 | 0.73 | 147.00 | 0.44 |
| 19.89 | 24.57 | 0.81 | 11.82 | 16.14 | 0.74 | 147.00 | 0.44 |
| 19.93 | 24.60 | 0.81 | 11.71 | 16.12 | 0.73 | 145.19 | 0.43 |
| 19.92 | 24.57 | 0.81 | 11.79 | 16.20 | 0.73 | 145.19 | 0.43 |
| 19.94 | 24.62 | 0.81 | 11.80 | 16.17 | 0.73 | 145.19 | 0.43 |
| 19.92 | 24.63 | 0.81 | 11.86 | 16.19 | 0.74 | 145.19 | 0.44 |
| 19.93 | 24.53 | 0.82 | 11.84 | 16.17 | 0.74 | 145.19 | 0.44 |
| 20.12 | 24.57 | 0.82 | 11.25 | 15.79 | 0.72 | 143.37 | 0.40 |
| 20.00 | 24.60 | 0.82 | 11.74 | 15.92 | 0.74 | 141.56 | 0.42 |
| 20.03 | 24.59 | 0.82 | 11.83 | 16.10 | 0.74 | 147.00 | 0.44 |
| 20.03 | 24.46 | 0.82 | 11.85 | 16.06 | 0.74 | 145.19 | 0.44 |